\documentclass[lettersize,journal]{IEEEtran}

\usepackage{cite}
\usepackage{algpseudocode}
\usepackage{bm}
\usepackage{amsfonts}
\usepackage{graphicx}
\usepackage{textcomp}
\usepackage{amsmath}
\usepackage{xcolor}
\usepackage{booktabs}
\usepackage{multirow}
\usepackage{setspace}
\usepackage{stfloats}
\usepackage{algorithm}
\usepackage{array}
\usepackage{amssymb}
\usepackage{color}
\usepackage{soul}
\usepackage{threeparttable,flushend}
\usepackage[caption=false,font=normalsize,labelfont=sf,textfont=sf]{subfig}
\usepackage{url}
\usepackage{verbatim}
\usepackage{tikz}

\begin{document}

\title{Hybrid Robotic Grasping with a Soft Multimodal Gripper and a Deep Multistage Learning Scheme\\}

\author{Fukang Liu, Fuchun Sun,~\IEEEmembership{Fellow,~IEEE,} Bin Fang,~\IEEEmembership{Member,~IEEE,} Xiang Li,~\IEEEmembership{Senior Member,~IEEE,} Songyu Sun,\\ and Huaping Liu,~\IEEEmembership{Senior Member,~IEEE}
\thanks{This work was supported in part by the Major Project of the New Generation of Artificial Intelligence of China (Grant No. 2018AAA0102900) and in part by the National Natural Science Foundation of China (Grant No. 62173197), Tsinghua University (Department of Computer Science and Technology)-Siemens Ltd., China Joint Research Center
for Industrial Intelligence and Internet of Things. 
\textit{(Corresponding authors:Bin Fang; Fuchun Sun.)}}
\thanks{Fukang Liu is with Carnegie Mellon University, Pittsburgh, PA 15213 USA (email: fukangl@andrew.cmu.edu).}
\thanks{Fuchun Sun, Bin Fang and Huaping Liu are with the Department of Computer Science and Technology, Tsinghua University, Beijing 100084, China (email:  fcsun@mail.tsinghua.edu.cn; fangbin@tsinghua.edu.cn; hpliu@tsinghua.edu.cn).}
\thanks{Xiang Li is with the Department of Automation, Tsinghua University, Beijing 100084, China (email: xiangli@tsinghua.edu.cn).}
\thanks{Songyu Sun is with the University of California, Los Angeles, CA 90095 USA (email: songyusun@g.ucla.edu).}
}

\maketitle

\begin{abstract}
Grasping has long been considered an important and practical task in robotic manipulation. Yet achieving robust and efficient grasps of diverse objects is challenging, since it involves gripper design, perception, control and learning, etc. Recent learning-based approaches have shown excellent performance in grasping a variety of novel objects. However, these methods either are typically limited to one single grasping mode or else more end effectors are needed to grasp various objects. In addition, gripper design and learning methods are commonly developed separately, which may not adequately explore the ability of a multimodal gripper. In this paper, we present a deep reinforcement learning (DRL) framework to achieve multistage hybrid robotic grasping with a new soft multimodal gripper. A soft gripper with three grasping modes (i.e., \textit{enveloping}, \textit{sucking}, and \textit{enveloping\_then\_sucking}) can both deal with objects of different shapes and grasp more than one object simultaneously. We propose a novel hybrid grasping method integrated with the multimodal gripper to optimize the number of grasping actions. We evaluate the DRL framework under different scenarios (i.e., with different ratios of objects of two grasp types). The proposed algorithm is shown to reduce the number of grasping actions (i.e., enlarge the grasping efficiency, with maximum values of $161.0\%$ in simulations and $153.5\%$ in real-world experiments) compared to single grasping modes.
\end{abstract}

\section{Introduction}
\IEEEPARstart{D}{exterous} manipulation is one of the primary goals in robotics\cite{b1}. Designing robotic hands and achieving flexible grasping are important and challenging tasks in object manipulation. Although significant progress has been made in gripper design and in robot learning targeted at robotic grasping, there still exist great limitations in achieving efficient and effective grasping for objects with different features (e.g., different weights, sizes, shapes, and textures) simultaneously with a single gripper.

The robotic hand is the key component for grasping. Traditional rigid grippers rely on precise control, which requires sophisticated sensing devices and closed-loop force systems\cite{b2} to grasp objects safely. In addition, because they lack adaptability, they are not capable of grasping a wide range of objects with different physical properties, and their hard contacts make them unsafe for interactions in a dynamic human environment\cite{b3,b4}. In recent years, soft grippers have elicited considerable attention because of advantages such as their lightweight, low cost, injury-free capability, and high adaptability\cite{b5}. For example, a soft gripper can adapt to objects with various geometric surface structures or stiffnesses to achieve conformal contacts\cite{b6,b7,b8}. Grippers made of flexible and soft components also overcome the inability of a rigid gripper to achieve soft contacts and safe interactions \cite{b8,b9,b10}. Nevertheless, most current soft grippers have only one single grasping mode, which causes difficulty in grasping objects with the different geometric and physical features needed to fulfill various robot tasks\cite{b11}. Although great progress has been made in the mechanical design and robust control of such grippers, they have limited ability to grasp and manipulate objects autonomously due to the lack of a combination of grasping approaches.

\begin{figure*}[htbp]
\centerline{\includegraphics[scale=0.13]{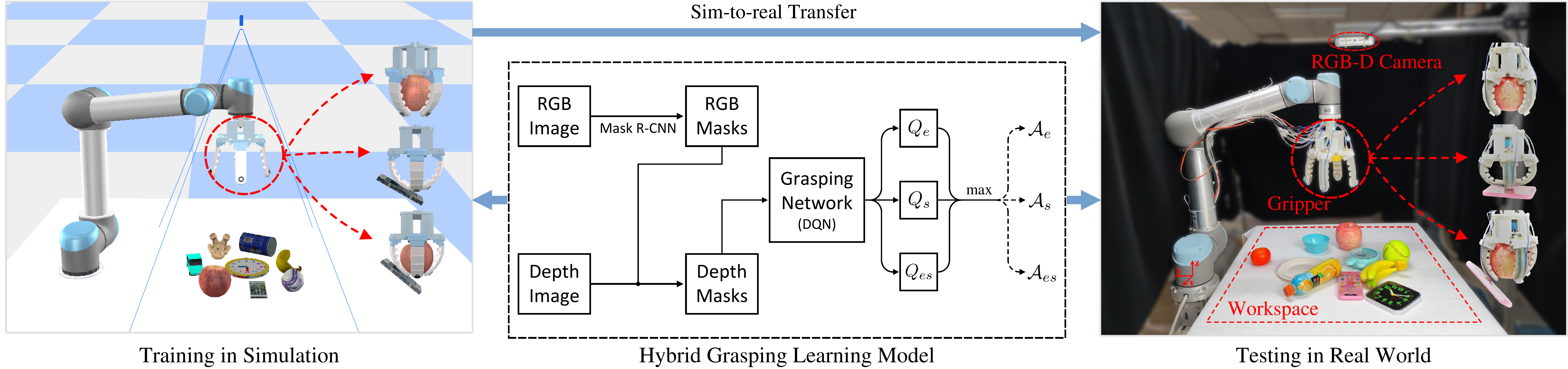}}
\caption{ System overview. The multimodal gripper employs three grasping modes—{\em enveloping}, {\em sucking} and {\em enveloping\_then\_sucking}—which can grasp two objects simultaneously. The hybrid grasping learning model acquires RGB-D images (captured by an overhead camera) as input and outputs three Q-value matrices. The robot executes the action with the highest value. The DRL algorithm is trained in simulations and then transferred into real-world robots.}
\label{fig:overview}
\vspace{-0.3cm}
\end{figure*}

In general, robotic grasping methodologies can be categorized into model-based grasping approaches and learning-based methods\cite{b12,b13}. The former use information about the 3D geometry of an object, and they require a perception system to recognize the target objects. However, they are not robust when confronted with a variety of novel objects or in dynamic environments\cite{b14,b15,b16}. The latter methods are based on machine learning and use training models to predict the probability and guide robotic grasping. State-of-the-art learning-based methods can handle different scenarios and achieve effective grasps with high accuracy and robustness. However, the majority of current works not only need to acquire large-scale training data to learn a good grasp but also face the challenge of poor generalizability. Learning grasping policies through simulations can alleviate the problem of data collection, but the simulation-to-real-world (sim-to-real) transfer still remains challenging\cite{b13,b17,b18,b19}. Bridging the gap between simulation and reality is thus key to accomplishing more efficient transfer. Furthermore, although various learning methods have been proposed to tackle the grasping problem, existing algorithms mainly focus on one single grasping mode (e.g., grasping, pushing, and suction)\cite{b20,b21}. Even though some previous studies have considered multimodal grasping\cite{b22}, grasping modes are commonly trained in an isolated manner (i.e., only a single mode is activated at a time), where the ability to grasp more than one object simultaneously with one gripper is not fully explored. Therefore, the problem of hybrid robotic grasping, where multiple grasping modes work simultaneously, has not been addressed systematically.

In this work, we develop a framework to achieve hybrid robotic grasping, including {\em gripper design}, {\em grasping modeling}, and {\em policy training}, as shown in Fig.~\ref{fig:overview}, and the proposed framework can generate a sequence of multimodal grasping actions to optimize the efficiency of object grasping. First, we design a soft multimodal gripper (SMG), capable of simultaneously grasping multiple objects with different geometrical and physical properties. On the one hand, the designed gripper can realize multimodal grasping modes and it can grab and release multiple objects at once with the help of vacuum actuators. On the other hand, because it is equipped with a layer-jamming variable-stiffness structure, the gripper not only can provide sufficient grasping force but also can perform robust grasping with great adaptability.

Second, we propose an end-to-end model based on deep reinforcement learning (DRL) to describe the hybrid grasping skills. Our value-based (Q-learning) method suits the SMG well and can fully explore its ability by minimizing the number of grasping actions. It takes depth images as inputs and outputs three Q-value matrices for actions to be executed for multimodal grasping. The robot takes the action with the highest Q-value to interact with the target objects.

Third, in order to overcome the limitations of gathering real-world data, we train our DRL model in the simulation environment of CoppeliaSim\cite{b24}. A simplified theoretical model is built into the simulation, which can precisely simulate the grasping process of the designed soft gripper in the real world. In addition, our DRL algorithm uses only depth images as inputs. Since the proposed framework only receives information about the 3D shape of an object (i.e., without color), training in the simulation environment is relatively efficient. The aforementioned measures guarantee the feasible sim-to-real transfer of the DRL policy.

In summary, the article presents the following contributions:
\begin{enumerate}
\item We have designed an SMG that can grasp a variety of objects with different physical properties. It is able to achieve robust grasping with the advantages of high compliance, strong stability, and precise control. Furthermore, with its three grasping modes, our soft gripper extends the current grasping capability to hybrid grasping, where multiple heterogeneous objects are grasped simultaneously.
\item We propose a hybrid grasping model based on DRL. A well-trained model can describe the variable structure of the SMG in an end-to-end manner, and hence such a model can be used to explore the capability of the SMG fully, that is, by enabling the grasping of multiple different objects simultaneously with a minimal number of steps.
\item We present several experiments in both simulation and real-world environments to evaluate the performance of our proposed method. The results demonstrate that our model exhibits high performance in minimizing the number of grasping actions in scenarios with different ratios of two types of objects. Furthermore, our DRL policy—trained in simulations exhibits good generalization to the real world. To the best of our knowledge, this is the first work to achieve hybrid robotic grasping with the combination of a multimodal gripper and  a multistage robot-learning model based on DRL. 
\end{enumerate}

The rest of this paper is organized as follows. Section~\ref{sec:related_work} briefly reviews related work in soft-gripper design and vision-based grasping. Section~\ref{sec:gripper_design} presents the basic idea of our SMG. Section~\ref{sec:method} explains the DRL framework used for multimodal grasping in this article. The simulations and real-world experiments are presented in Section~\ref{sec:results}. Finally, Section~\ref{sec:conclusion} concludes this article and the Appendix presents the configurations of the gripper in the action space.

\section{Related Work}
\label{sec:related_work}
\subsection{Soft Gripper}
To compensate for the shortcomings of traditional rigid grippers, soft and extensible materials are increasingly being studied for the design of lighter, simpler, and more universal grippers\cite{b5}. Soft grippers are usually classified by actuation into two major categories\cite{b4,b5,b7,b10}. In the first class, the grippers are driven by external motors to achieve soft grasping. This makes it possible to obtain a high grasping force easily by choosing suitable motors since they are external to and independent from the gripper\cite{b5}. There are two main types of externally actuated soft grippers: contact-driven deformation grippers\cite{b25,b26,b27} and tendon-driven grippers\cite{b6,b29,b30}. In the second class, the soft grippers are the actuators themselves. They are mainly of three types: pneumatic soft actuators\cite{b31,b32,b33,b34}, shape-memory-alloy actuators\cite{b35,b36,b37} and electroactive polymers\cite{b38,b39}. The wide use of these actuators based on fluid dynamics and smart materials enables more flexible and intelligent soft robotic grippers. In addition, in order to tackle the challenge of the poor load capacity of a soft gripper, various designs of grippers with variable stiffness have been proposed. Generally, these approaches either use layer/granular/fiber jamming \cite{b8,b41,b42} or else employ materials with controllable stiffness (e.g., shape-memory materials and low-melting-point alloys)\cite{b39,b43} to obtain variable stiffness. Furthermore, inspired by soft biological structures, suction cups have been integrated into soft grippers for the purpose of lifting flat objects and enhancing grasp stability\cite{b7,b34,b44}. 

However, the previous works have two major drawbacks: 1) Most grippers are limited to one single grasping mode, and 2) the methods are limited to mechanical design and lack learning algorithms to improve grasping and manipulation capabilities. In this work, we designed a new SMG that utilizes both a layer-jamming structure and a tendon-driven mechanism to achieve relatively high stiffness and grasping force. The SMG can perform three grasping modes in total: \textit{enveloping}, \textit{sucking} and \textit{enveloping\_then\_sucking}. The last is a combined mode that can grasp two objects at the same time.

\subsection{Learning-Based Robotic Grasping}
Learning-based approaches to robotic grasping have been widely studied and applied in robotics. Advancements in machine learning have enabled significant progress in robotic grasping. Machine vision can provide comprehensive information about the target objects and the environment, enabling the robot to achieve a robust grasp. Many vision-based grasping methods use visual features obtained from RGB or RGB-D images for object detection and grasp determination\cite{b45,b46,b47}. Methods using only RGB, or both RGB and depth information, can improve the performance of grasping by making pixel-level grasp detection possible\cite{b20,b21,b22,b48}, but they are relatively sensitive to color information and have poor adaptability to novel objects. Some works plan grasps based on only depth images or use only RGB images that are invariant to the object color for object detection and recognition\cite{b50,b51,b52}, which enhances the grasping robustness.

Training a learning-based grasping algorithm depends heavily on collecting large-scale datasets. Some related works use supervised-learning approaches and train grasping based on human-annotated data\cite{b45,b54,b55,b56}. However, human labeling is costly and time-consuming, and artificial errors may influence the quality of the labels. In contrast, DRL techniques have been increasingly adopted to enable robotic grasping to learn policies through trial-and-error processes. These techniques include policy-gradient methods\cite{b12,b53}, deep Q-learning methods\cite{b20,b57,b58} and actor-critic algorithms\cite{b59,b61}. Furthermore, In order to alleviate the data collection problem, simulation environments are often utilized for training both the supervised and the DRL algorithms\cite{b19,b53,b58}. However, the sim-to-real transfer remains challenging for current robotic-learning methods. Bridging the sim-to-real gap and accomplishing more efficient policy transfer is a crucial step toward achieving a robust grasp. 

In parallel, several results have been reported in the literature for the development of the robotic grasping system with multiple end effectors in order to facilitate the grasping of various objects with different geometric and physical properties. For example, by integrating a gripper with two or three fingers and a suction cup, such a system can execute both suction and parallel-jaw grasps to manipulate varied objects\cite{b13,b22,b62}. For example, Kengo \textit{et al.}\cite{b63} have presented a gripper consisting of three fingers with a suction mechanism at each fingertip, which enables some dexterous manipulations. Shun \textit{et al.}\cite{b21} have proposed a multimodal grasping fusion system, which can simultaneously execute suction and a pinch grasp. However, limited by the design of the grippers or the grasping policies, these works handle the grasping modes separately. Only one grasping action can be applied to one object at a time. 

In an effort to overcome the limitations of the previous work, In this paper, we propose a new learning-based framework to achieve hybrid robotic grasping. Specifically, our developed SMG which has a variable structure and high compliance is able to envelop or suck objects in a mixed manner, providing the foundation for hybrid grasping. The proposed multistage learning scheme can generate a sequential optimal grasping strategy, subject to the unknown and nonlinear model of the SMG. Thus,  the hardware and the software complement each other, which guarantees the minimum number of grasping actions on multiple heterogeneous objects (i.e., maximal grasping efficiency).

\section{Gripper Design}
\label{sec:gripper_design}
\begin{figure*}[htbp]
\centerline{\includegraphics[scale=0.165]{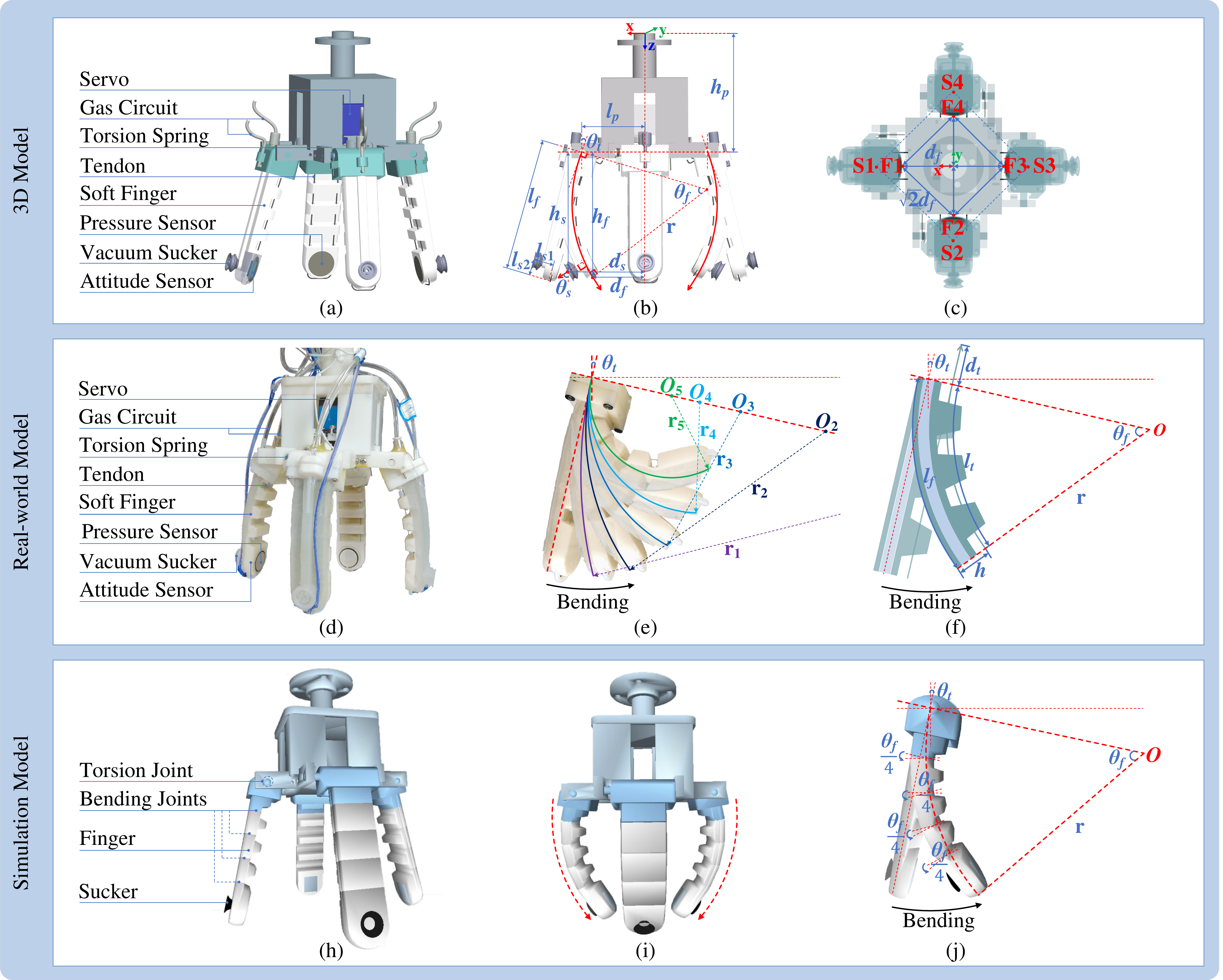}}
\caption{\label{fig:gripperdesign} 3D model and prototype of the SMG. (a) 3D model of the gripper and its main components. (b) Front view of the gripper. (c) Top view of the gripper. The items labeled F1–F4 and S1–S4 are the positions of the four fingertips and four suckers of the bent gripper, respectively. (d) Real-world model of the gripper and its main components. (e) Different bending shapes of a finger for different displacements of the tendon. The deformed shapes correspond well to circular arcs. The items labeled $O_1$–$O_5$ and $r_1$–$r_5$ are the centers and radii of the bending shapes of the finger, respectively. (f) The theoretical model of a finger showing its bent shape. (h) Simulation  model of the gripper and its main components. (i) Bent state of a rigid-body gripper. (j) Bending deformations of the finger can be achieved by using four joint angles.}
\vspace{-0.3cm}
\end{figure*}

In this paper, we propose an improved SMG based on our previous work\cite{b11}. As shown in Fig.~\ref{fig:gripperdesign}, this gripper is designed to be symmetrical with four fingers evenly distributed around the vertical axis (the direction of gravity), with each finger consisting of a soft chamber, jamming layers, force sensors, and a sucker [see Fig.~\ref{fig:gripperdesign}(a)]. The bending of each finger can be achieved by the dragging of tendons, which are driven by the servo. We can adjust the limiting size of the gripper opening by using torsion springs with different leg angles. In addition, vacuum suckers installed on the backs of the fingertips are used to generate suction. All of the air channels are independent, although they share a single vacuum source, guaranteeing stable suction forces. Moreover, each fingertip is equipped with a piezoresistive pressure sensor to obtain the contact forces and an attitude sensor to acquire pose information from the finger. Such a design ensures precise control and allows the SMG to grasp multiple heterogeneous objects in a hybrid manner.

\subsection{Kinematic Model of the SMG}
In order to calculate the precise positions of the suckers and fingertips, a local coordinate system $G$ is established on the pedestal of the gripper, as shown in Figs.~\ref{fig:gripperdesign}(b) and (c). We assume that the shape of a bent finger is a circular arc. The vertical distance $h_{f}$ between the fingertip and the torsion spring can be written in the form
\begin{equation}
\label{equ:hf}
\begin{aligned}
\begin{split}
h_{f} &= r[\operatorname{sin}(\theta_t)+\operatorname{sin}(\theta_f-\theta_t)]
\end{split}
\end{aligned}
\end{equation}
where $r$, $\theta_t$ and $\theta_f$ are the bending radius of the finger, the angle between the axis of the finger and the vertical direction, and the bending angle of the finger, respectively.

The distance $d_{f}$ between the fingertip and the \textit{z}-axis of coordinate frame $G$ can be written as
\begin{equation}
\label{equ:df}
\begin{aligned}
\begin{split}
d_{f} &= l_{p}-r[\operatorname{cos}(\theta_t)-\operatorname{cos}(\theta_f-\theta_t)]
\end{split}
\end{aligned}
\end{equation}
where $l_p$ denotes the distance between the torsion spring and the \textit{z}-axis of frame $G$.

Similarly, the vertical distance $h_s$ between the sucker and the torsion spring can be written as
\begin{equation}
\label{equ:hs}
\begin{aligned}
\begin{split}
h_{s} = \;& r[\operatorname{sin}(\theta_t)+\operatorname{sin}(\theta_f-\theta_t)]\\
& -l_{s1}\operatorname{sin}(\theta_s)+l_{s2}\operatorname{cos}(\theta_s)
\end{split}
\end{aligned}
\end{equation}
where $\theta_s$ is the angle between the direction normal to the sucker and the \textit{z}-axis of frame $G$, and $l_{s1}$ and $l_{s2}$ represent the vertical and horizontal distances between the sucker and the fingertip, respectively.

The distance $d_s$ between the sucker and the \textit{z}-axis of frame $G$ is
\begin{equation}
\label{equ:ds}
\begin{aligned}
& \begin{split}
d_{s} = \;& l_{p}-r[\operatorname{cos}(\theta_t)-\operatorname{cos}(\theta_f-\theta_t)]\\
& +l_{s1}\operatorname{cos}(\theta_s)+l_{s2}\operatorname{sin}(\theta_s).
\end{split}
\end{aligned}
\end{equation}

Thus, the coordinates of the suckers and fingertips of each of the four fingers relative to frame $G$ can be obtained, respectively, as
\begin{equation}
\label{equ:xs}
\begin{aligned}
\left(\bm{x_s},\bm{y_s},\bm{z_s}\right)=
\left[\begin{tabular}{ccc}
$d_s$ & 0 & $h_{p}+h_{s}$ \\
0 & $-d_s$ & $h_{p}+h_{s}$ \\ 
$-d_s$ & 0 & $h_{p}+h_{s}$ \\
0 & $d_s$ & $h_{p}+h_{s}$ \\
\end{tabular}\right]
\end{aligned}
\end{equation}

\begin{equation}
\label{equ:xf}
\begin{aligned}
\left(\bm{x_f},\bm{y_f},\bm{z_f}\right)=
\left[\begin{tabular}{ccc}
$d_f$ & 0 & $h_{p}+h_{f}$ \\
0 & $-d_f$ & $h_{p}+h_{f}$ \\ 
$-d_f$ & 0 & $h_{p}+h_{f}$ \\
0 & $d_f$ & $h_{p}+h_{f}$ \\
\end{tabular}\right]
\end{aligned}
\end{equation}
where $h_p$ denotes the \textit{z}-coordinate value of the torsion spring in frame $G$. Each row in (\ref{equ:xs}) corresponds sequentially to S1 through S4, and each row in (\ref{equ:xf}) to F1 through F4. To grasp objects of different sizes, the \emph{opening distance} of the gripper must be adjusted appropriately. We define the opening distance $d$ as the distance between two adjacent fingertips [see Fig.~\ref{fig:gripperdesign}(c)]:
\begin{equation}
\label{equ:d}
\begin{aligned}
d = \sqrt{2}d_f.
\end{aligned}
\end{equation}

Given the value of $d$, we can solve for $\theta_f$ (i.e., the angular measure of the finger in the bent state) by using (\ref{equ:df}) and (\ref{equ:d}) and send it to the gripper as a control command.

\begin{table}[htbp]\footnotesize
    \caption{\label{tab:gripperdata} Real-world Gripper Model Parameters}
    \begin{center}
    \footnotesize
        \begin{tabular}{p{1.9cm} p{2.3cm} l}
            \toprule
            \textbf{Component}&\textbf{Symbol}&\textbf{Value}\\
            \midrule
            \multirow{2}*{\textbf{Fishing Lines}}& Max Pulling Force &$21.8$ kg\\
                                                 & Wire Diameter &$0.32$ mm\\
            \midrule
            \multirow{3}*{\textbf{Paper Sheet}}& Size &$0.10\times24\times110$ $(\rm{mm})$\\
                                                & Young's Modulus &$6$ GPa\\
                                                & Poisson's Ratio &$0.2$\\
            \midrule
            \multirow{3}*{\textbf{Steel Sheet}}& Size &$0.10\times24\times110$ $(\rm{mm})$\\
                                                & Young's Modulus &$20.6$ GPa\\
                                                & Poisson's Ratio &$0.3$\\
             \midrule
            \multirow{2}*{\textbf{\shortstack{Attitude Sensor\\(JY61P)}}}& Size &$15.24\times15.24\times2$ $(\rm{mm})$\\
                                                & Angle Accuracy &\textit{x}-, \textit{y}-axes: $0.05^{\circ}$,  \textit{z}-axis: $1^{\circ}$\\
            \midrule
            \multirow{2}*{\textbf{\shortstack{Pressure Sensor\\(D2027)}}}& Measuring Range &$0.5\sim100$ $(\rm{kg})$\\
                                                & Sensing Diameter &$20$ (mm)\\
            \midrule
            \textbf{Ecoflex 0050}& Hardness &$50$ (shore 00)\\
            \midrule
            \textbf{Nasil 4230} &Hardness &$28$ (shore A)\\
            \bottomrule
            \end{tabular}
        \label{bs}
    \end{center}
\vspace{-0.4cm}
\end{table}

\subsection{Prototype of the SMG}
We constructed a prototype of the SMG by referring to the 3D model, depicted in Figs.~\ref{fig:gripperdesign}(d)–(f). We fabricated the soft actuator of the finger using two types of silicone rubber: the softer Ecoflex 0050 and the harder Nasil 4230. We used flexible fishing lines as U-shaped tendons to drive the four fingers and achieve different modes of bending deformation. In addition, we stacked the jamming layers—which consisted of $18$ layers of paper and spring-steel sheets—together layer by layer. We then stuffed them inside the soft chamber for the purpose of achieving effective variable stiffness in order to improve the multimodal grasping performance. Increasing the air pressure in the chamber mainly stiffens the finger without producing much bending deformation.
Table~\ref{tab:gripperdata} lists the main parameters of the finger components.

\begin{figure}[htbp]
\centerline{\includegraphics[scale=0.15]{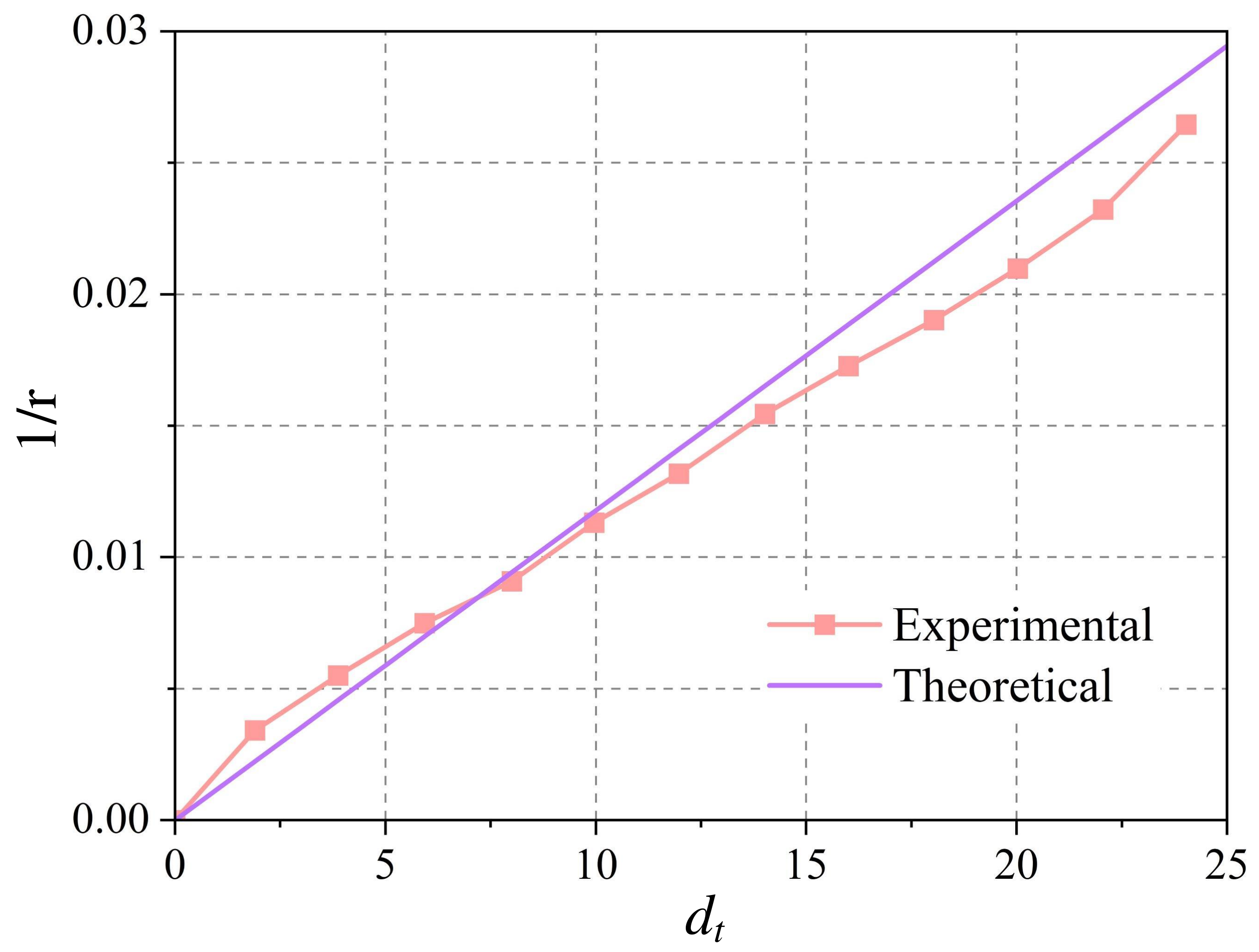}}
\caption{\label{fig:gripperdeformation} Bending shapes of the finger under different displacements of the tendon line. Real-world experiments and theoretical model are shown in Figs.~\ref{fig:gripperdesign}(e) and (f), respectively.}
\vspace{-0.3cm}
\end{figure}

Before enveloping or sucking an object, the four soft fingers are first driven by the tendons to achieve given bending angles and deformations. In order to ensure precise control of the fingers, it is necessary to define the relationship between the displacement of the tendon and the bending shape of the finger. Fig.~\ref{fig:gripperdesign}(e) shows a snapshot of a series of experiments illustrating different bending shapes of a finger for different displacements of the tendon. We found that the deformed shapes correspond well to circular arcs, which verifies the assumption in the 3D model that the bending shape is a circular arc.

For the sake of implementation, we constructed a simplified deformation model to describe the relationship between the bending angle $\theta_f$ of the finger and the displacement $d_t$ of the tendon line, as shown in Fig.~\ref{fig:gripperdesign}(f), where $l_t$ and \textit{h} denote the length of the tendon line in the finger and the perpendicular distance from the tendon line to the back of the finger, respectively. The  parameters in the simplified deformation model are then given by
\begin{equation}
\label{equ:lf}
\begin{aligned}
& l_f = r\theta_f\\
& l_t = (r-h)\theta_f\\
& l_f = l_t + d_t
\end{aligned}
\end{equation}
where $l_f$ is the length of the finger. From (\ref{equ:lf}), the bending angle can be expressed as
\begin{equation}
\label{equ:thetaf}
\begin{aligned}
&\theta_f=\frac{d_t}{h}.
\end{aligned}
\end{equation}

Substituting (\ref{equ:thetaf}) and (\ref{equ:d}) into (\ref{equ:df}) yields

\begin{equation}
\label{equ:d_2}
\begin{aligned}
\begin{split}
\frac{d}{\sqrt{2}}=l_{p}-\frac{l_{f}h}{d_t}[\operatorname{cos}(\theta_t)-\operatorname{cos}(\frac{d_t}{h}-\theta_t)].
\end{split}
\end{aligned}
\end{equation}

Hence, for a desired opening distance $d$, the value of the displacement $d_t$ of the tendon line can be obtained from (\ref{equ:d_2}), which is considered as the control input in real-world experiments. From (\ref{equ:lf}), we can obtain the curvature of the bent finger:
\begin{equation}
\label{equ:curvature}
\begin{aligned}
&\frac{1}{r}=\frac{d_t}{hl_f}.
\end{aligned}
\end{equation}

Fig. 3 depicts the kinematic relationship between the displacement of the tendon line and the geometry curvature of the finger. It indicates that the actual test results are close to the theoretical results.
 
We also established a simulation environment to train the DRL model of the SMG. As shown in  Figs.~\ref{fig:gripperdesign}(h)–(j), each finger consists of four joints with identical joint angles that enable to achieve bending deformations of circular shapes [see in Fig.~\ref{fig:gripperdesign}(j)].

\section{Method}
\label{sec:method}
The grasping process involves object detection, grasp selection and robot (gripper) control. Because of the variable structure and the deformable property of the SMG, it is very difficult to develop an exact model to capture all of its features. In addition, due to its high active/passive DOFs and constraints, planning and control of the SMG also are not trivial.

Finding a policy to minimize the number of grasping actions is inherently a sequential-learning problem, where the SMG interacts with an uncertain grasping environment, and its actions may affect future situations. Therefore, in this work, we used reinforcement learning to generate multimodal grasping actions that guarantee optimal grasping efficiency.

In this section, we develop an end-to-end DRL framework for hybrid grasping with the developed SMG, including {\em modeling}, {\em policy search}, and {\em training \& implementation}. The proposed framework explores the advantages of the SMG and facilitates the grasping of multiple heterogeneous objects at the same time.

\subsection{Problem Formulation}
In general, a hybrid grasping task can be described as a Markov decision process (MDP): given a state $S_t$ at time \textit{t}, a robot chooses and executes an action $A_t$ under a policy $\pi$($S_t$) \cite{b69}. As a consequence of its action, the robot transitions to a new state $S_{t+1}$ and receives a reward $R_{t+1}$. Reinforcement learning is learning an optimal policy in order to maximize the expected discounted return:
\begin{equation}
\label{equ:gt}
G_{t} = \sum_{k=0}^{n}\gamma^{k}R_{t+k+1}
\end{equation}
where $\gamma$ is the discount rate, $0\leq\gamma\leq1$.

 Mnih \emph{et al.}\cite{b64} developed a Deep Q-network, which combines Q-learning with a deep convolutional neural network (CNN). In this work, we use a Double Deep Q-learning (DDQN) algorithm\cite{b65} to obtain the optimal policy $\pi_{*}$ that maximizes the action-value function $\operatorname{Q}(S_t,A_t) = \sum\limits_{k=0}^{n}\gamma^{k}R_{t+k+1}$. In DDQN, finding an optimal policy (or the optimal value function) means minimizing the temporal-difference error $\delta_{t}$, which can be written as
\begin{equation}
\label{equ:delta_t}
\delta_{t} = \left| Y_{t}-Q(S_t,A_t;\bm{\theta_t)}\right|
\end{equation}
\begin{equation}
\label{equ:yt}
Y_{t} = R_{t+1}+\gamma\operatorname{Q}\left(S_{t+1},\arg\max_{a}\operatorname{Q}(S_{t+1},a;\bm{\theta_t}),\bm{\theta_{t}^{-}}\right)
\end{equation}
where $\bm{\theta}$ and $\bm{\theta^{-}}$ are the parameters of the online network and the target network, respectively, and $\delta_{t}$ denotes the difference between the estimated and the desired value of $S_t$.

In an MDP, the agent interacts continually with the environment. Actions are taken by the agent, and the environment then responds to these actions and presents new situations to the agent. In this work, the robot selects grasping actions at each of a sequence of discrete time steps under a given policy. We define the state $\mathcal{S}$, the action $\mathcal{A}$, the reward $\mathcal{R}$, our Q-function framework and the training details depending upon the specific hybrid grasping task; that is, they are defined to achieve the simultaneous grasping of multiple objects at the same time.

\subsection{States}
\begin{figure}[htbp]
\centerline{\includegraphics[scale=0.126]{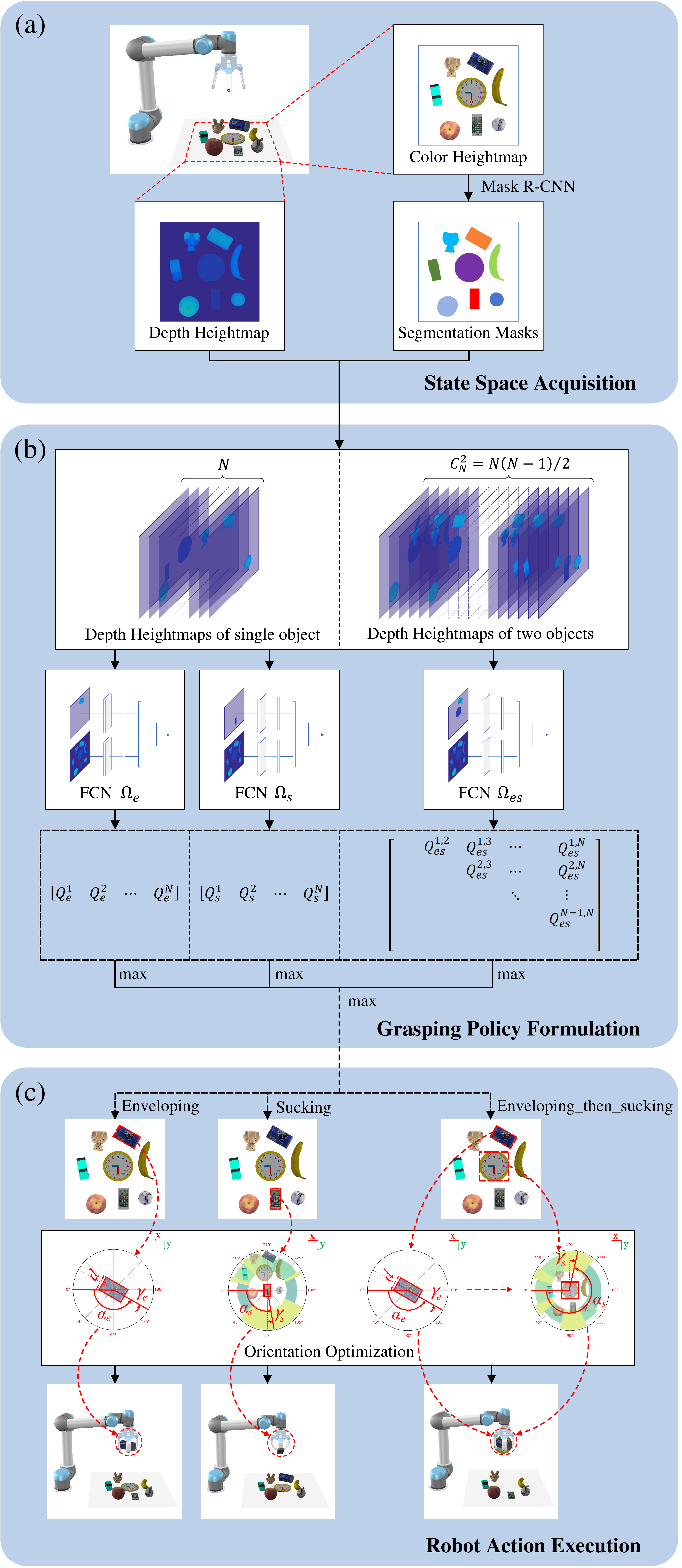}}
\caption{\label{fig:algorithm} Learning multimodal grasping policies for the SMG. (a) Color and depth heightmaps are generated through projection transformations of the RGB-D images captured by the depth camera. The edges of the heightmaps are predefined with respect to the boundaries of the robot's workspace. masks are then obtained by using Mask R-CNN. Finally, we get the depth masks. (b) Three Deep Q-networks takes as inputs the depth heightmaps of one or two objects (i.e., local properties) concatenated with the depth heightmap for all the objects (i.e., global properties), and they output three Q-value matrices. (c) The action that maximizes the Q-value is executed. For an \textit{enveloping} action, enveloping grasping is applied to the target object by bending the four fingers. For a \textit{sucking} action, one sucker executes sucking onto the target object. An \textit{enveloping\_then\_sucking} action is a combination of enveloping and sucking executed on two target objects, separately and sequentially.}
\end{figure}

\begin{figure*}[htbp]
\centerline{\includegraphics[scale=0.14]{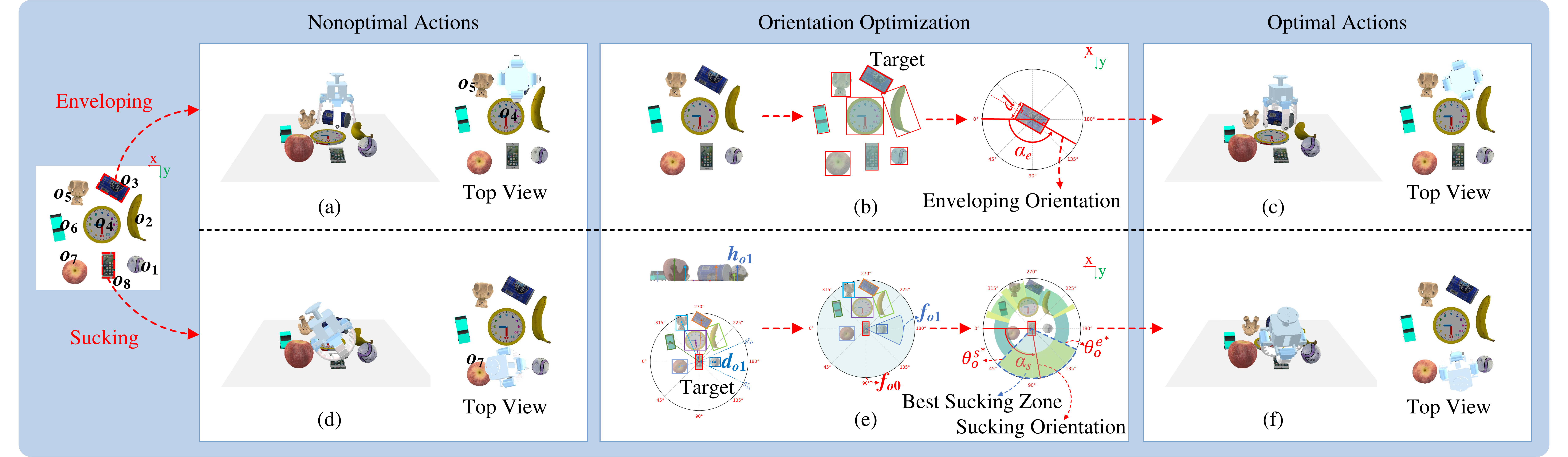}}
\caption{\label{fig:optimization} Determination of the orientations for enveloping and sucking. (a) A nonoptimal \textit{enveloping} action involving collisions between the gripper and nontarget objects (i.e., $O_4$ and $O_5$). (b) Determination of the enveloping orientation and opening distance for the target object. Segmentation masks are used to determine the minimum-area rotated rectangle. The enveloping orientation $\alpha_e$ denotes the angle between the long axis of the rectangle and the \textit{x}-axis of the gripper. The opening distance $d$ denotes the short side of the rectangle. (c) An optimal \textit{enveloping} action without collisions. (d) A nonoptimal \textit{sucking} action involving a collision between the gripper and a nontarget object $O_7$. (e) Determination of the sucking orientation. (Left) The maximum height of each object and the distance between the centers of each nontarget object and of the target object are used to determine the obstacle factor $f_{oi}$, and minimum-area bounding boxes are used to obtain the obstacle regions of nontarget objects. (Center) The obstacle factor for each object is obtained. (Right) Finally, the best sucking zone and sucking orientation are determined for the target object based on the obstacle factors of all the objects in the entire workspace. (f) An optimal \textit{sucking} action without collisions.}
\vspace{-0.3cm}
\end{figure*}

We define the state space $\mathcal{S}$ as a series of R-GBD images captured by a depth camera. These RGB-D images contain both color and depth information for all the objects in the environment. We use the workspace UR5 robot from Universal Robots\cite{b72}, which covers an area of $0.448\times0.448$ $\rm{m^2}$ in our experiments. First, each RGB-D image is preprocessed by using a projection transformation to yield both color and depth heightmaps, the edges of which are predefined with respect to the boundaries of the robot's workspace. The heightmaps are then rendered with a pixel resolution of $224\times224$, as illustrated in Fig.~\ref{fig:algorithm}(a). Second, we apply Mask R-CNN\cite{b23} to the RGB heightmap images to obtain segmentation masks for each object. The depth heightmap for each object can then be obtained. Further, based on the segmentation masks, we use minAreaRect OpenCV to find the minimum-area rotated rectangle (i.e., a bounding box) for each object, the location of which can be defined as the center of the bounding box, as shown in Fig.~\ref{fig:optimization}(b).

Let $\mathcal{O}=(\mathcal{O}_1,\mathcal{O}_2,\cdots,\mathcal{O}_{N})$ denote the state of the environment at time step \textit{t}. The quantity $\mathcal{O}_i=(\mathcal{M}_i^C,\mathcal{M}_i^D,\mathcal{M}_i^B)$ specifies the information about object \textit{i}, i.e., its color $\mathcal{M}_i^C$, depth $\mathcal{M}_i^D$, and bounding box $\mathcal{M}_i^B$. The proposed states quickly capture key geometrical information about the object, which determines the subsequent grasping mode.

\begin{figure*}[htbp]
\centerline{\includegraphics[scale=0.13]{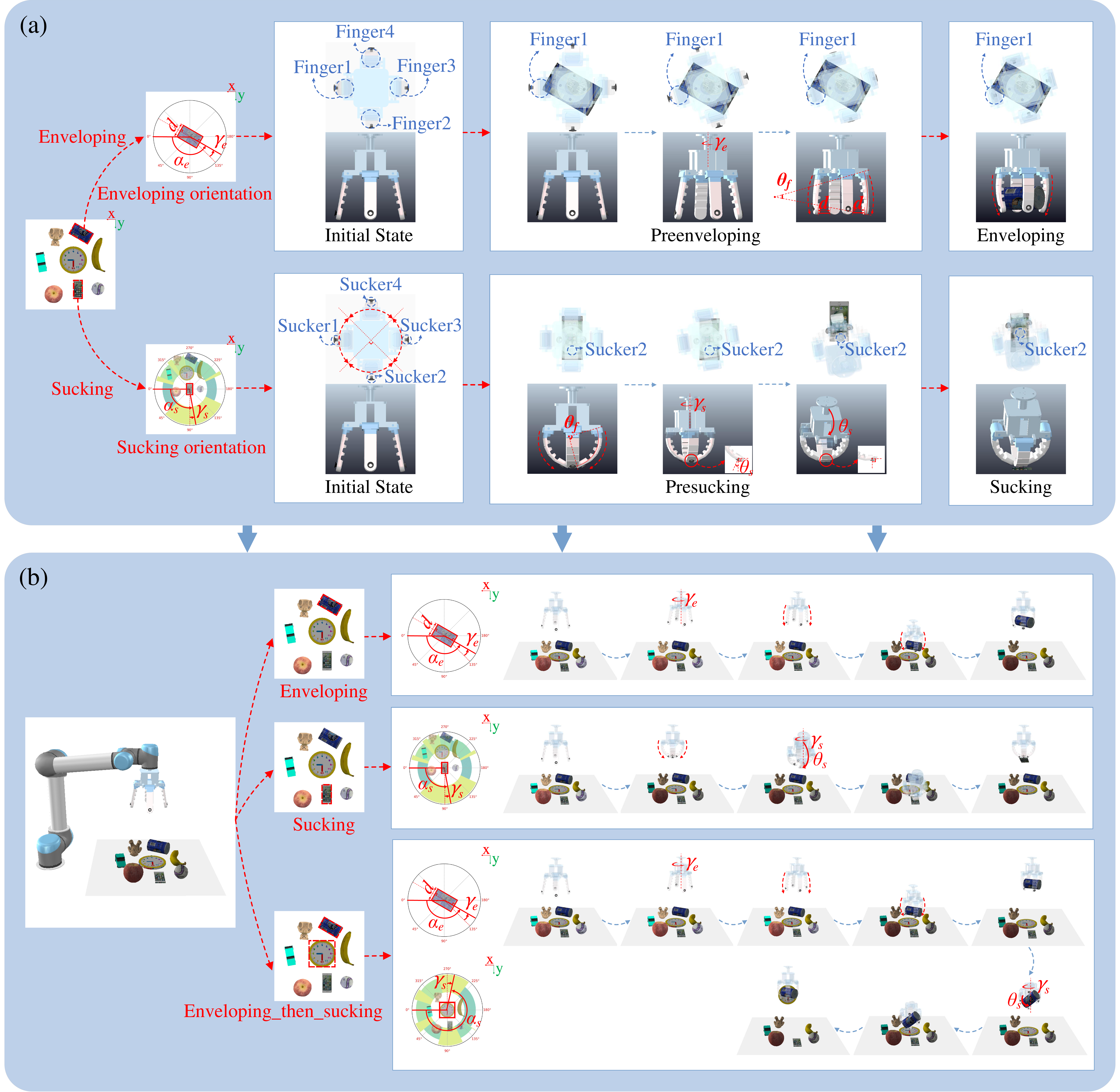}}
\caption{\label{fig:action} Three grasping actions: \textit{enveloping}, \textit{sucking}, and \textit{enveloping\_then\_sucking}. (a) Enveloping and sucking poses of the gripper. To envelop a target object, the enveloping orientation $\alpha_e$, enveloping rotation angle $\gamma_e$, and gripper opening distance $d$ must first be determined. A pre-enveloping process is then followed, including a rotation through the angle $\gamma_e$ around the $z$-axis and a bending displacement of the fingers to achieve the enveloping opening distance $d$. Finally, the gripper approaches the target object, and a further bending displacement is applied to the fingers to achieve an enveloping action. For sucking, the sucking orientation $\alpha_s$ and the sucking rotation angle $\gamma_s$ must first be derived. Then a pre-sucking process is executed, including a bending displacement of the fingers, followed by a rotation through the angle $\gamma_s$ around the $z$-axis and then a rotation through the angle $\theta_s$ around the $x$- axis of frame $G$. Finally, sucker2 approaches the target position and executes a sucking action. (b) The processes involved in each of the three actions in the 3D grasping environment. The \textit{enveloping\_then\_sucking} action is a combination of an \textit{enveloping} action and a \textit{sucking} action.}
\vspace{-0.3cm}
\end{figure*}

\subsection{Grasping Actions}
The action space $\mathcal{A}$ denotes the grasping behavior in 3D space defined by the configuration of the gripper, $ T=(p,R)\in\rm{SE(3)}$, where $p\in\mathbb{R}^3$ specifies the position of the gripper from the origin of the robot’s coordinate frame $W$, and $R\in \rm{SO(3)}$ specifies the orientation of the gripper relative to frame $W$. The space $\mathcal{A}$ includes three sub-action spaces—\textit{enveloping} $\mathcal{A}_{e}$; \textit{sucking} $\mathcal{A}_{s}$; and \textit{enveloping\_then\_sucking} $\mathcal{A}_{es}$—which are respectively parametrized by $ T_e=(p_e,R_e)$, $T_s=(p_s,R_s)$ and $T_{es}=(p_e,R_e)\cup(p_s,R_s)$. These three subaction spaces are defined as follows:

\noindent\textbf{Enveloping:} The notation $\mathcal{A}_{e}$ denotes an enveloping grasp executed at the target position $q_e$ in frame $W$, which specifies the 3D location of the center of the bounding box of the target object.

A nonoptimal grasping orientation or a larger opening of the gripper may create undesirable collisions between the gripper and nontarget objects, which may contribute to unsuccessful grasping [see Fig.~\ref{fig:optimization}(a)]. In order to achieve robust enveloping without collisions [see Fig.~\ref{fig:optimization}(c)], we aim to optimize the enveloping orientation and opening distance of the gripper. As shown in Fig.~\ref{fig:optimization}(b), the axis for enveloping is along the longer side of the bounding box, just as human hands grasp objects. The enveloping orientation $\alpha_e$ ($0^{\circ}\leq\alpha_e<180^{\circ}$) denotes the angle between the enveloping axis and the \textit{x}-axis of frame $W$, and $d$ denotes the length of the shorter side of the rectangle, which is the same as the opening distance between two adjacent fingertips. In addition, our gripper contains four identical fingers, and they overlap completely with a minimum rotation angle $90^{\circ}$ around the vertical axis (\textit{z}-axis). Therefore, as illustrated in Fig.~\ref{fig:action}(a), a minimum angle of rotation, $\gamma_e$, can be determined with respect to the enveloping orientation:
\begin{equation}
\label{equ:gamma_e}
{\gamma_e} =
\begin{cases}
\alpha_e-45^{\circ},&{\text{if}}\ \alpha_e\leq90^{\circ}\\
\alpha_e-135^{\circ}, &{\text{otherwise.}}
\end{cases}
\end{equation}

As shown in Fig.~\ref{fig:action}(b), a complete \emph{enveloping} process is as follows:

\begin{enumerate}
\item
The gripper moves to a location directly above the target object.
\item
The gripper rotates through an angle $\gamma_e$  around the vertical direction, to reach the enveloping orientation $\alpha_e$. 
\item
A given bending angle $\theta_f$ is applied to each of the four fingers to achieve an opening distance $d$. The bending is achieved by pulling the tendon line in the real world or by rotating the torsional joints in the simulations, as discussed in Section~\ref{sec:gripper_design}. This preenveloping process minimizes collisions between the gripper and non-target objects to the maximum extent possible. 
\item
The gripper moves to the target position and further bending is applied to the fingers, finally achieving an enveloping action.
\end{enumerate}

\noindent\textbf{Sucking:} The notation $\mathcal{A}_{s}$ denotes suction executed at the target position $q_s$ in frame $W$. When performing a tilted grasp on a target object in dense and cluttered environment, a robot needs to determine a grasp orientation that avoid nearby objects. A random sucking orientation may result in collisions between the gripper and nontarget objects [see Fig.~\ref{fig:optimization}(d)]. To secure collision-free sucking [see Fig.~\ref{fig:optimization}(f)], the sucking orientation of the gripper has to be optimized.

Both the geometry and position of the obstacles (or nontarget objects) around the target object influence the determination of the sucking orientation. On the one hand, an obstacle of smaller width contributes to a larger range for the selection of sucking orientation; on the other hand, an obstacle with a smaller height or a larger distance between the obstacle and the target object leads to a larger action area for side grasping, and vice versa. We introduce three specific parameters to describe the \textit{\textbf{obstacle factors}} $f_{oi}$ for objects around the gripper; they are $\mathcal{M}_i^B$, the bounding box for the $i^{th}$ object; $h_{oi}$, the height of the $i^{th}$ object; and $d_{oi}$, which is the distance between the centers of the $i^{th}$ object and the target object. An illustration with obstacles of different sizes is shown in Fig.~\ref{fig:optimization}(e).  

We define the obstacle factor $f_{oi}$ as
\begin{equation}
\label{equ:f_oi}
f_{oi} =
\begin{cases}
{\text{exp}}(-\frac{h_{oi}-h_{o0}}{d_{oi}}),&{\text{if}}\ h_{oi}>h_{o0}\\
1, &{\text{otherwise}}
\end{cases}
\end{equation}
where $h_{o0}$ denotes the height of the target object. Thus, a larger $d_{oi}$ or a smaller $h_{oi}$ produces a larger $f_{oi}$, which means a better-sucking orientation, and vice versa.

For each nontarget object, we can obtain six sector regions by connecting the center point of the target object to any two points in the \textit{four-point set} of the bounding box for this nontarget object. Then, we define the largest sector region $\theta_{oi}$ among the six as its obstacle area:
\begin{equation}
\label{equ:theta_oi}
\mathcal\theta_{oi} =
\begin{cases}
(\theta_{oi}^s,\theta_{oi}^e),&{\text{if}}\ \theta_{oi}^s < \theta_{oi}^e\\
(0,\theta_{oi}^e) \cup (\theta_{oi}^s,360^{\circ}),&{\text{otherwise}}
\end{cases}
\end{equation}
where $\theta_{oi}^s$ and $\theta_{oi}^e$ denote the starting angle and the ending angle, respectively, for $\theta_{oi}$, as shown in Fig.~\ref{fig:optimization}(e). Therefore, the angle-dependent obstacle factor for the $i^{th}$ object can be written as
\begin{equation}
\label{equ:ff_oi}
\mathcal{F}_{oi}(\theta) =
\begin{cases}
f_{oi},&{\text{if}}\ \theta \in \theta_{oi}\\
1,&{\text{otherwise}}
\end{cases}
\end{equation}
and the angle-dependent obstacle factor for all the objects is defined as
\begin{equation}
\label{equ:f_o}
\begin{aligned}
\mathcal{F}_{o}(\theta) = \prod_{i=0}^{N}\mathcal{F}_{oi}(\theta).
\end{aligned}
\end{equation}

Those value of $\mathcal{F}_{o}$ in the 2D plane is illustrated in Fig.~\ref{fig:optimization}(e). The \textit{best sucking zone} corresponds to the continuous space that has both the largest $\mathcal{F}_{o}$ and largest angular extent. The Appendix describes the selection of the sucking orientation.

Algorithm 2 (see the Appendix) determines the sucking orientation $\alpha_s$ ($0^{\circ}\leq\alpha_s<360^{\circ}$), which can be expressed as
\begin{equation}
\label{equ:alpha_s}
\alpha_s =
\begin{cases}
\theta_P,&{\text{if}}\ \theta_o^{s^{\ast}}\leq \theta_o^{e^{\ast}}\\
\theta_P-180^{\circ},&{\text{if}}\ (\theta_o^{s^{\ast}}> \theta_o^{e^{\ast}}\ {\text{and}}\ \theta_P\geq 180^{\circ})\\
\theta_P+180^{\circ},&{\text{otherwise}}
\end{cases}
\end{equation}
where $\theta_P=(\theta_o^{s^{\ast}}+\theta_o^{e^{\ast}})/2$. The quantities $\theta_o^{s^{\ast}}$ and $\theta_o^{e^{\ast}}$ denote the starting angle and the ending angle, respectively, of the \textit{best sucking zone}. 

Our gripper contains four suckers, one on the back of each fingertip. In order to make one sucker—$S_i$, where $i=1, 2, 3$ or 4—achieve the sucking orientation $\alpha_s$, the gripper must rotate through the minimum angle $\gamma_s$ around the vertical direction, as illustrated in Fig.~\ref{fig:action}(a). For each sucker $i$, the angle $\gamma_s$ can be written as follows:
\begin{equation}
\label{equ:gamma_s}
\begin{aligned}
\gamma_s=\left\{
\begin{array}{lll} 
\alpha_s,&i=1,& {\text{if}}\;\alpha_s\leq 45^{\circ}\\
\alpha_s-90^{\circ},&i=2,& {\text{if}}\;45^{\circ}<\alpha_s\leq 135^{\circ}\\
\alpha_s-180^{\circ},&i=3,& {\text{if}}\;135^{\circ}<\alpha_s\leq 225^{\circ}\\
\alpha_s-270^{\circ},&i=4,& {\text{if}}\;225^{\circ}<\alpha_s\leq 315^{\circ}\\
\alpha_s-360^{\circ},&i=1, & {\text{otherwise}}.
\end{array}
\right.
\end{aligned}
\end{equation}

As shown in Fig.~\ref{fig:action}(b), a complete \emph{sucking} process is as follows:
\begin{enumerate}
\item 
A given bending angle $\theta_f$ is applied to each of the four fingers to achieve a bending state, while the gripper moves to the location directly above the target object. 

\item
The gripper rotates through an angle $\gamma_s$ around the vertical (gravity) direction, such that the sucker reaches the sucking orientation $\alpha_s$.

\item
While maintaining the angle $\gamma_s$, the gripper rotates further through the angle $\theta_s$ around the axis (the \textit{x}-axis or \textit{y}-axis of frame $G$) perpendicular to the sucking orientation, so that the direction normal to the sucker is vertical (i.e., parallel to the direction of gravity). Note that the angle $\theta_s$ is obtained with the attitude sensor in the real world or by the dynamic module in the simulations. 

\item
The sucker moves to the target suction point and executes a sucking action.
\end{enumerate}

\noindent\textbf{Enveloping\_then\_sucking:} The notation $\mathcal{A}_{es}$ denotes a combination of an \textit{enveloping} action executed at the target position $q_e$ and a \textit{sucking} action executed at the target position $q_s$. During an enveloping\_then\_sucking attempt, the robot needs to grasp two objects sequentially. 

A complete \emph{enveloping\_then\_sucking} process is as follows [see Fig.~\ref{fig:action}(b)]:

\begin{enumerate}
\item
The gripper moves to location directly above the target object to be enveloped and rotates through an angle $\gamma_e$ around the vertical direction, reaching the enveloping orientation $\alpha_e$. 
\item
A given bending angle $\theta_f$ is applied to each of the four fingers to achieve the opening distance $d$. 
\item
The gripper moves to the target position $q_e$ and further bending is applied to the fingers, finally achieving an enveloping action.
\item 
The gripper next moves to a location directly above the target object for sucking and rotates through the angle $\gamma_s$ around the vertical (gravity) direction, so that the sucker reaches the sucking orientation $\alpha_s$.
\item
While maintaining the angle $\gamma_s$, the gripper rotates further through an angle $\theta_s$ around the axis (the \textit{x}-axis or \textit{y}-axis of frame $G$) perpendicular to the sucking orientation, so that the direction normal to the sucker is vertical (i.e., parallel to the direction of gravity). 
\item
The sucker moves to the target suction point $q_s$ and executes a sucking action.
\end{enumerate}

These actions actually provide the motion primitives for hybrid grasping, and the combination of those actions enables the SMG to deal with multiple heterogeneous objects simultaneously.

\subsection{Rewards}
In this paper, the problem of hybrid grasping is formulated under the DRL framework, so that the objective of the developed SMG is to maximize the rewards; that is, to minimize the number of grasping actions needed in order to reduce the total running time necessary to pick up all the objects in the workspace. To fulfill this requirement, the rewards in our algorithm are designed as follows:

The state of rewards $\mathcal{R}$ has a finite number of elements; we define $\mathcal{R} = \left\{2.5,1,0.5,0 \right\}$. For an \textit{enveloping} or \textit{sucking} action, the reward $R_e(s_t,s_{t+1})$ or $R_s(s_t,s_{t+1})$ equals $1$ for a successful grasp and $0$ otherwise. For an \textit{enveloping\_then\_sucking} action, $R_{es}(s_t,s_{t+1})=2.5$ for a fully successful action that picks up two objects, $0.5$ for a semi-successful action that only picks up one object, and $0$ otherwise. Piezoresistive pressure sensors and gas-pressure sensors are used to verify the \textit{enveloping} and \textit{sucking} actions in real-world experiments.

The reward value for executing a fully successful \textit{enveloping\_then\_sucking} action is larger than the sum of the rewards for executing one successful \textit{enveloping} and one successful \textit{sucking} action separately. This encourages the robot to perform as many \textit{enveloping\_then\_sucking} actions as possible and to minimize the number of grasping actions when there are two types of objects for grasping (i.e., those suitable for enveloping and those suitable for sucking). Further, the reward value for executing a semi-successful \textit{enveloping\_then\_sucking} action is smaller than the reward for executing either a successful \textit{enveloping} or \textit{sucking} action. Consequently, we expect the robot to take the latter actions when there is only one type of object in the workspace (i.e., one suitable for enveloping or one suitable for sucking). This is reasonable, since ${\rm max}\{t_e, t_s\}\leq t_{es}\leq t_e+t_s$, where $t_e, t_s$, and $t_{es}$ denote the running times for complete \textit{enveloping}, \textit{sucking}, and \textit{enveloping\_then\_sucking} processes, respectively, and the reward setting guides the robot to minimize the running time necessary to pick up all the objects.

\subsection{Learning Functions}
Traditional grasping methods are usually trained in an isolated manner, and only one single grasping mode is activated at a time. However, the difficulties of finding good ways to represent the properties of states and optimization algorithms to achieve multimodal grasping raise a significant challenge for multistage hybrid grasping. In this work, we used Mask R-CNN to separate local properties of states (i.e., the depth heightmaps of one or two objects) from their global properties (i.e., the depth heightmap of all the objects). We then develop a DDQN framework to process these heightmap images, as shown in Fig.~\ref{fig:algorithm}. Our Q-function networks consist of three fully convolutional networks (FCNs)—$\varOmega_e$, $\varOmega_s$, and $\varOmega_{es}$—which represent three grasp modes: \textit{enveloping}, \textit{sucking}, and \textit{enveloping\_then\_sucking}, respectively. Our hybrid grasping model takes a sequence of multimodal actions that leads to the state with the highest expected future reward in order to achieve optimal grasping efficiency. Our Q-function networks are closely related to that of Zeng \textit{et al.}~\cite{b20}.

The three FCNs are constructed using the same network architecture. 
First, two parallel DenseNet-121~\cite{b66} towers pre-trained on ImageNet\cite{b67} are taken as the inputs to the depth channels for the global and local properties, as shown in Fig.~\ref{fig:algorithm}(a). The depth heightmaps are preprocessed by scaling, padding, and normalizing them. Second, Each DenseNet-121 layer is followed by channel-wise concatenation and one traditional convolutional block, which contains two convolutional layers, interleaved with batch normalization (BN) and ReLU activation functions: BN $\rightarrow$ ReLU $\rightarrow$ Conv ($1\times1$) $\rightarrow$ BN $\rightarrow$ ReLU $\rightarrow$ Conv ($20\times20$). Third, the Q-values $\left(\emph{i.e.}, Q_e, Q_s, Q_{es}\right)$ are obtained.

As shown in Fig.~\ref{fig:algorithm}(b), the inputs and outputs of the Q-function networks are different with respect to the three primitive actions. For the \textit{enveloping} action $\mathcal{A}_{e}$ and the \textit{sucking} action $\mathcal{A}_{s}$, we feed one object's depth heightmap image concatenated with the global depth image into $\varOmega_e$ or $\varOmega_s$ and obtain one Q-value for either enveloping or sucking to be applied to this object. If there are $N$ objects in state $S_t$, we obtain multiple Q-values for enveloping and sucking: $Q_e=\left\{Q_{e}^1\;Q_{e}^2\;\dots\;Q_{e}^N\right\}$ and $Q_s=\left\{Q_{s}^1\;Q_{s}^2\;\dots\;Q_{s}^N\right\}$, respectively. In contrast, if $N\textgreater1$, for the \textit{enveloping\_then\_sucking} action $\mathcal{A}_{es}$, we feed one depth heightmap image that contains two objects, concatenated with the global depth image into $\varOmega_{es}$, and get one Q-value for enveloping and sucking separately applied to the two objects. We then obtain $C_N^2=N(N-1)/2$ Q-values for enveloping\_then\_sucking

$$
Q_{es}= 
\left[\begin{tabular}{ccccc}\vspace{1ex}
\; & $Q_{es}^{1,2}$ & $Q_{es}^{1,3}$ &\dots & $Q_{es}^{1,N}$ \\
\; &\; & $Q_{es}^{2,3}$ &\dots & $Q_{es}^{2,N}$ \\ 
\; &\; & \;& $\ddots$ & $\vdots$ \\
\; &\; & \; & \; & $Q_{es}^{N-1,N}$ \\
\; &\; & \; & \; & \;\\
\end{tabular}\right]
$$

\begin{algorithm}[H] 
\caption{Action Selection}  
\label{alg:A}
\hspace*{0.02in}{\bf Input:} 
Q values: $Q_e, Q_s$ for $N=1$ and $Q_e, Q_s, Q_{es}$ for $N>1$\\
\hspace*{0.02in}{\bf Output:}
Maximum Q value $Q^{\ast}$, Action $\mathcal{A}^\ast$, Target object(s) index $n^\ast$

\begin{algorithmic}[1]
\State $Q_e^\ast \gets {\rm max}\;Q_{e},Q_s^\ast \gets {\rm max}\;Q_{s}$
\State $\displaystyle{n_e \gets \arg\max_{i}Q_{e}^i,n_s \gets \arg\max_{i}Q_{s}^i}$

\If{$N=1$}
    \If{$Q_e^\ast \geq {\rm max}\{Q_e^\ast,Q_s^\ast\}$}
        \State $Q^\ast \gets Q_e^\ast, \mathcal{A}^\ast \gets \textit{enveloping}, n^\ast \gets n_e$
    \Else
        \State $Q^\ast \gets Q_s^\ast, \mathcal{A}^\ast \gets \textit{sucking}, n^\ast \gets n_s$
    \EndIf
    
\Else
    \State $Q_{es}^\ast \gets {\rm max}\;Q_{es}$
    \State $\displaystyle{n_{es}^1,n_{es}^2 \gets \arg\max_{i,j}Q_{es}^{i,j}}$
    \If{$Q_e^\ast\geq{\rm max}\{Q_e^\ast, Q_s^\ast, Q_{es}^\ast\}$}
        \State $Q^\ast \gets Q_e^\ast, \mathcal{A}^\ast \gets \textit{enveloping}, n^\ast \gets n_e$
    \ElsIf{$Q_s^\ast\geq{\rm max}\{Q_e^\ast, Q_s^\ast, Q_{es}^\ast\}$}
        \State $Q^\ast \gets Q_s^\ast, \mathcal{A}^\ast \gets \textit{sucking}, n^\ast\gets n_s$
    \Else
        \State $Q^\ast \gets Q_{es}^\ast, \mathcal{A}^\ast \gets \textit{enveloping\_then\_sucking}$
        \If{$Q_e^{n_{es}^1}\geq Q_e^{n_{es}^2}$}
            \State $n^\ast \gets n_{es}^1$ (for \textit{enveloping}), $n_{es}^2$ (for \textit{sucking})
        \Else 
            \State $n^\ast \gets n_{es}^2$ (for \textit{enveloping}), $n_{es}^1$ (for \textit{sucking})
        \EndIf
    \EndIf
\EndIf
\State \bf{return:} $Q^{\ast}$, $\mathcal{A}^\ast$, $n^\ast$
\end{algorithmic} 
\vspace{-0.1cm}
\end{algorithm}

\begin{figure*}[htbp]
\centerline{\includegraphics[scale=0.44]{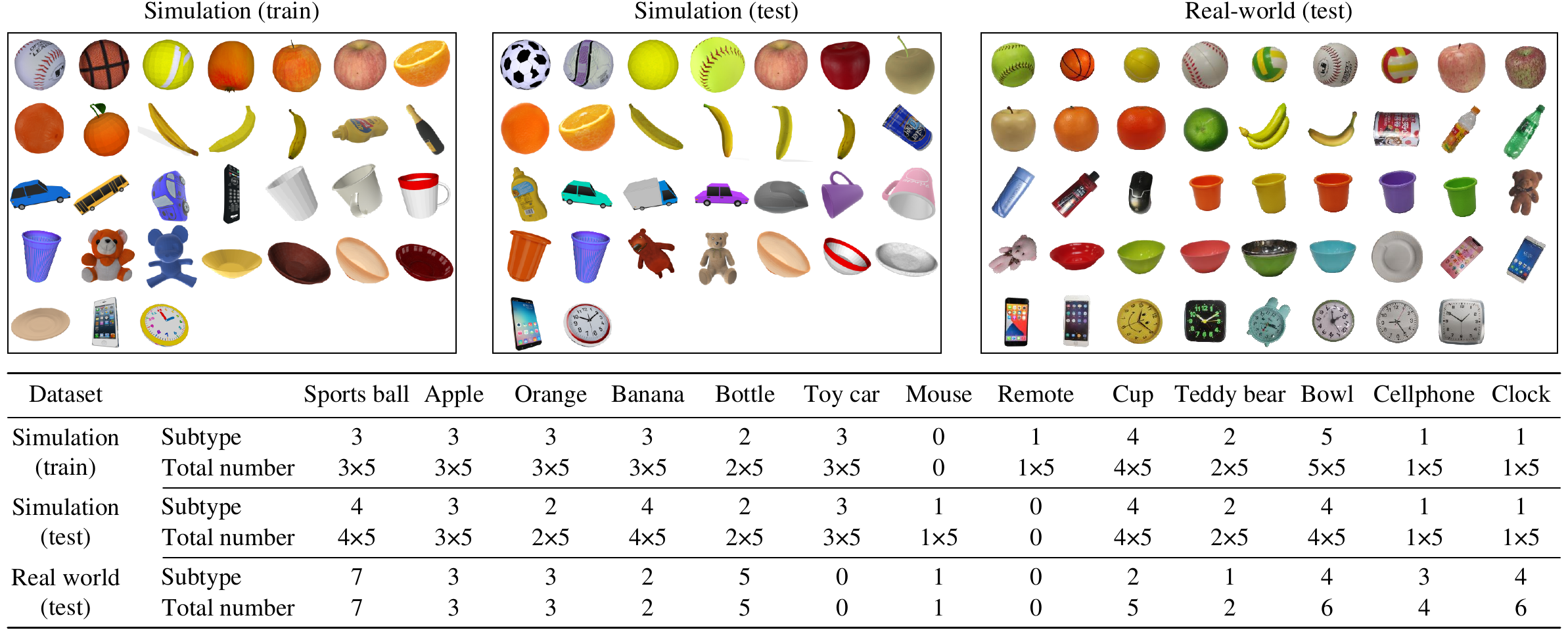}}
\caption{\label{fig:dataset} Datasets used for training and tests. Our datasets cover $13$ object categories, and each category contains one or more than one subtype. The simulation dataset is generated from a set of 3D simulation models, which are split randomly into training and test sets. We use the simulator CoppeliaSim to create five different sizes of objects for each subtype used in the simulations. The real datasets are collected from life scenes.}
\vspace{-0.3cm}
\end{figure*}

If there is a single object—that is, if $N=1$—the matrix $Q_{es}$ reduces to zero. As described in algorithm $1$, once we obtain $Q_e$, $Q_s$, and $Q_{es}$, we can determine the action to be executed. The action that maximizes the Q-function concerns the object(s) with the highest Q-value, i.e., $Q^\ast$. When $N=1$, the action $\mathcal{A}^\ast\in(\mathcal{A}_e,\mathcal{A}_s)$, is executed for one object; if $N>1$, $\mathcal{A}^\ast\in(\mathcal{A}_e,\mathcal{A}_s,\mathcal{A}_{es})$, and actions are applied to two objects when $\mathcal{A}^\ast=\mathcal{A}_{es}$, as shown in Fig.~\ref{fig:algorithm}(c).

\subsection{Training and Testing}
Our Q-function networks are trained by using the Adam optimizer \cite{b68} with the learning rate $10^{-4}$ and the Huber loss function:
\begin{equation}
\label{equ:loss}
\mathcal{L} =
\begin{cases}
\frac{1}{2}(\delta_{t})^2,&{\text{if}}\ \delta_{t}<1\\
\delta_{t}-\frac{1}{2},&{\text{otherwise}}
\end{cases}
\end{equation}
where the definition of $\delta_{t}$ is given in (\ref{equ:delta_t}).
Our Q-function networks share the same parameter values (e.g., step size, discount rate, exploration parameters). The network weights are reset using Kaiming normal initialization\cite{b70} before training. 
Our models are trained in PyTorch and the system uses an NVIDIA RTX 2080 Super and an Intel Core i7-10875H processor for computing. The exploration strategy is $\epsilon-$greedy, with $\epsilon$ initialized at $0.5$ and then annealed over training to $0.1$. Our future discount is constant at $0.5$\cite{b20}. 

During testing both in simulations and in the real world, we give the networks the same learning rate ($10^{-4}$) in order to prevent endless loops in which a given action occurs repeatedly while the state space remains unchanged. In addition, at the beginning of each new experiment, the network weights are set to their original state (after training and before testing).

\section{Results}
\label{sec:results}
In this section, we evaluate the performance of our SMG as well as the effectiveness of our proposed approach using a set of experiments. First, we introduce the training datasets. Second, we validate the multimodal grasping ability of our SMG and demonstrate the capabilities of our method to secure collision-free grasps. Then, we train our DRL algorithm and test its ability to adapt to multimodal grasping based on simulation environments. Finally, the accuracy and efficiency of our hybrid grasping framework are verified in both real-world and simulation experiments. The grasping efficiency is illustrated by using hybrid grasping of significantly different objects, e.g., flat and thin objects that are more suitable for sucking, and rounded objects that are more suitable for enveloping. The purpose of these experiments is clarified as follows:

\begin{enumerate}
\item To demonstrate that our designed soft gripper has good adaptability and flexibility in grasping different kinds of objects;
\item To evaluate the performance of our proposed algorithm and its ability to achieve multimode autonomous grasping; and
\item To study the influence of different ratios of two kinds of objects on grasping efficiency and to verify that our multimodal grasping mode is superior to one single grasping mode.
\end{enumerate}

\subsection{Datasets}
Our 3D object models are based on the COCO dataset\cite{b71}. In this work, we use Mask R-CNN for object detection and segmentation. Taking an RGB image as input, our COCO pre-trained Mask R-CNN model detects and segments objects above a threshold and then outputs segmentation masks and bounding boxes for all the objects in the image. 

In order to obtain 3D models for training and testing, we use both real-world data and synthetic data generated by computer simulations. Our real-world and simulation data each cover 11 and 13 object categories, respectively, with highly-varied geometries: \textit{sports ball, apple, orange, banana, bottle, toy car, mouse, remote, cup, teddy bear, bowl, cellphone}, and \textit{clock}. As shown in  Fig.~\ref{fig:dataset}, each object category is divided into one or more than one subtype according to the difference in geometry. In order to obtain sufficient objects for training and tests, every subtype used in the simulations contains five objects with random sizes, which are obtained from the CoppeliaSim robotics simulator\cite{b68}. All of the objects in the simulations are split randomly into training and test sets. Objects in the real-world dataset are collected from life scenes. Our simulation datasets used for training and tests contain 155 and 150 distinct object instances, respectively. Our real-world dataset contains 44 distinct objects in total. Furthermore, all of these 3D models can be roughly classified into two grasp types according to two grasping strategies: they are more suitable for enveloping (apples, oranges, bananas, etc.) or they are more suitable for sucking (clocks, cellphones, flat bottles, and bowls, etc.). We employ the two types of 3D objects in both simulations and real-world experiments.

\begin{figure*}[htbp]
\centerline{\includegraphics[scale=0.19]{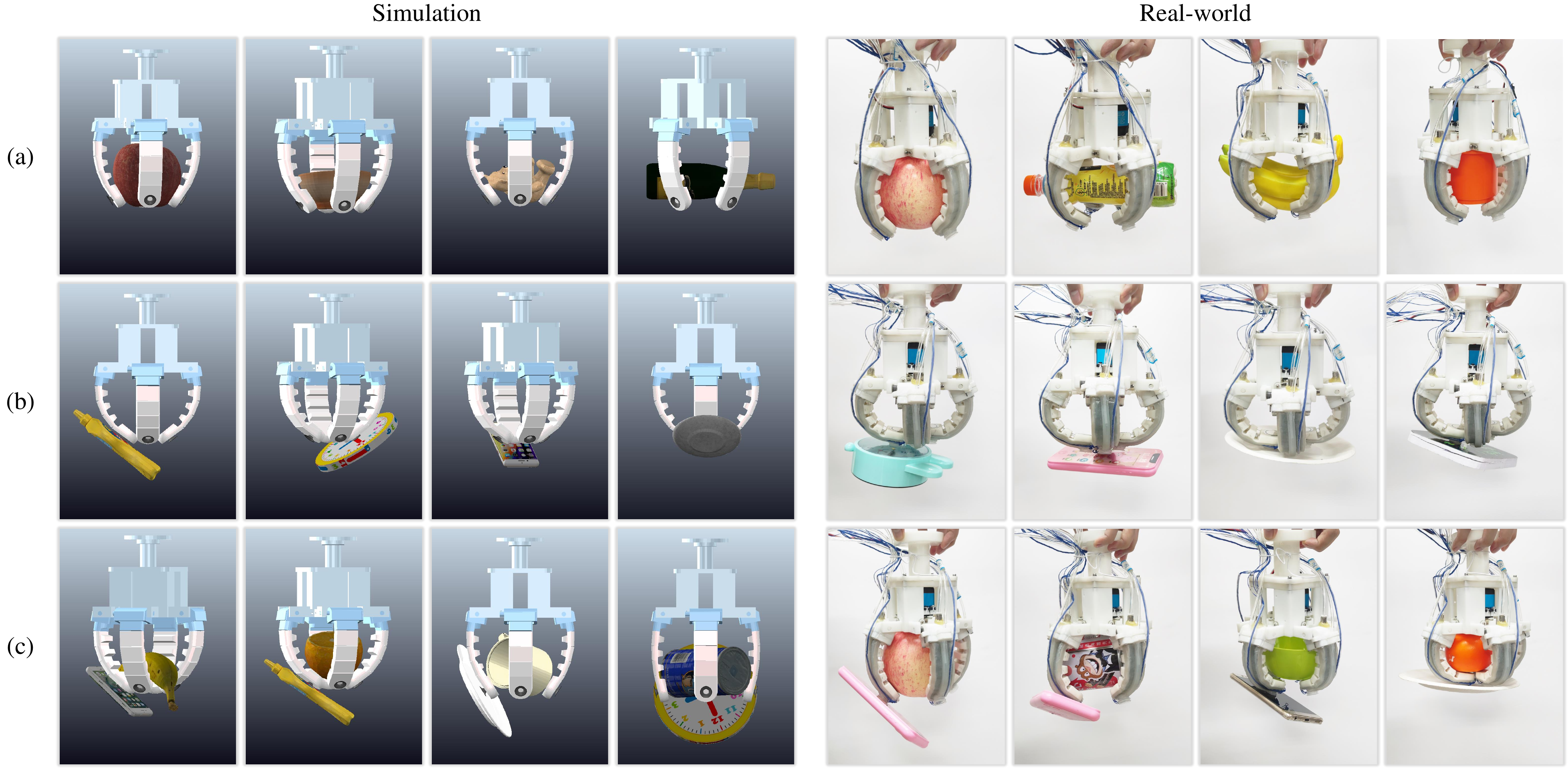}}
\caption{\label{fig:demenstration} Grasping demonstration of the multimodal gripper. (a) \textit{Enveloping} of objects with rounded shapes. (b) \textit{Sucking} of objects with flat surfaces. (c) \textit{Enveloping\_then\_sucking} of two types of objects.}
\vspace{-0.3cm}
\end{figure*}

\subsection{Prototype and Demonstration}
\textit{1) Multimodal Grasping:} We tested the multimodal grasping performance of the designed gripper using both simulations and real-world experiments. Fig.~\ref{fig:demenstration} shows that the gripper is able to grasp a variety of objects with different characteristics. Fig.~\ref{fig:demenstration}(a) presents the grasping of rounded objects, such as fruits, bottles, and cups. Fig.~\ref{fig:demenstration}(b) shows the sucking of flat objects, such as clocks, cellphones, and bottles with flat surfaces. Fig.~\ref{fig:demenstration}(c) shows the combined grasping of two types of objects (i.e., rounded and flat). These experiments demonstrate that our gripper has good grasping ability and adaptability.

\begin{figure*}[htbp]
\centerline{\includegraphics[scale=0.28]{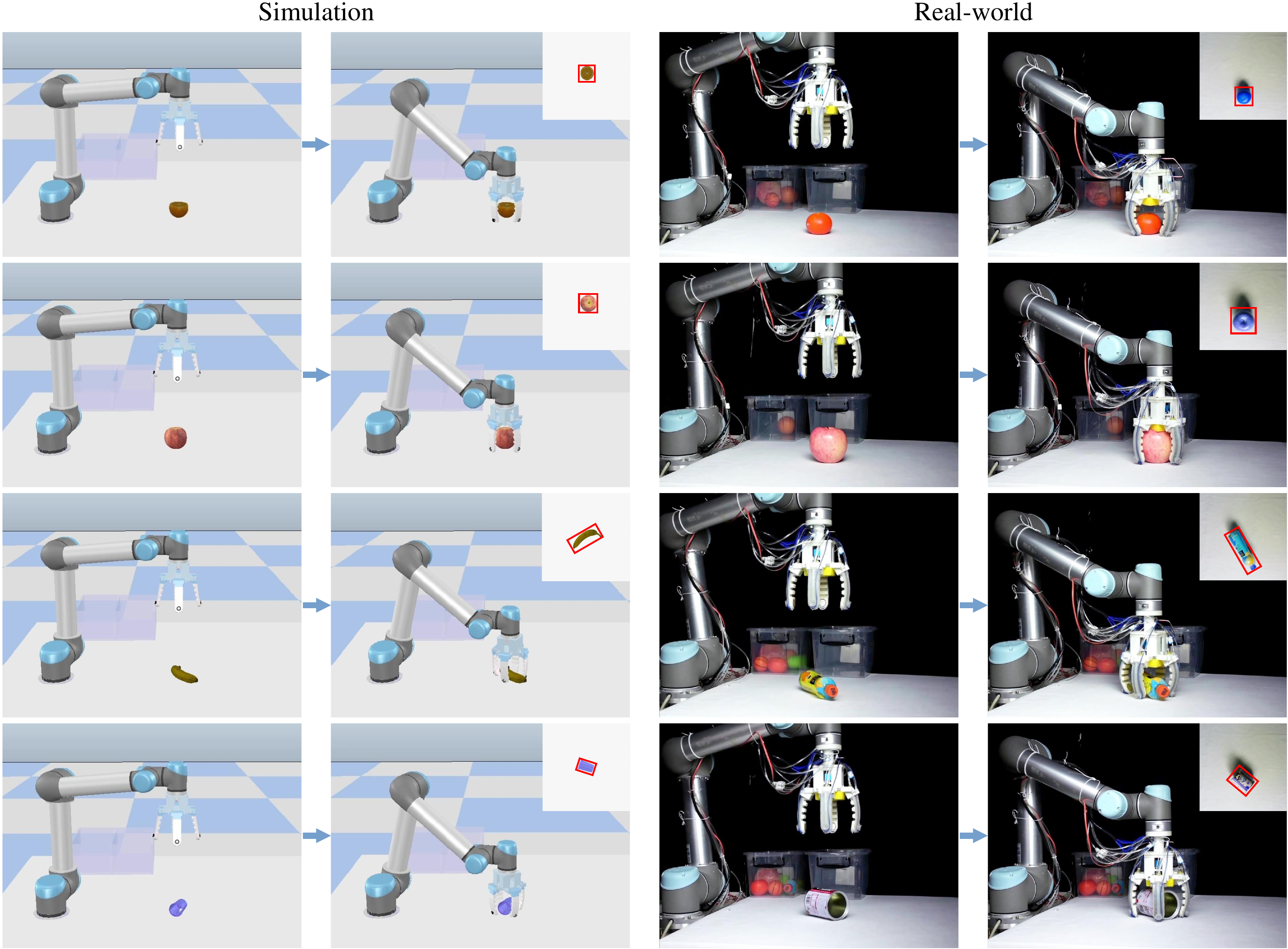}}
\caption{\label{fig:enveloping} Enveloping orientation and opening distance optimization for the four cases both in simulation and in real-world scenarios. Snapshots on the top-right corner of the images are the color heightmaps of each state. Target objects are surrounded by red rectangles, which are the minimum-area bounding boxes.}
\end{figure*}

\begin{figure*}[htbp]
\centerline{\includegraphics[scale=0.28]{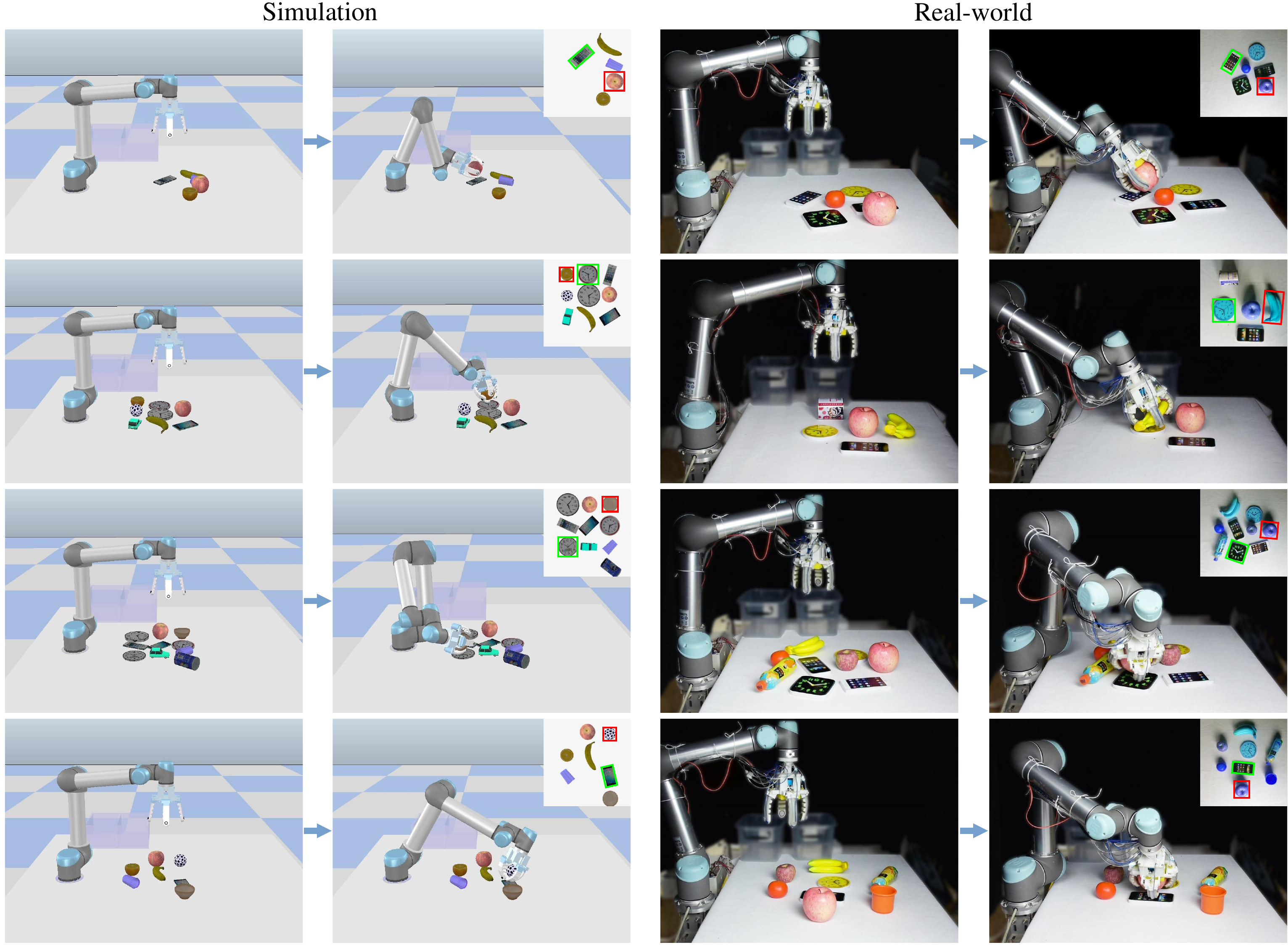}}
\caption{\label{fig:sucking} Sucking orientation optimization for the four cases both in simulation and in real-world scenarios. Snapshots on the top-right corner of the images are the color heightmaps of each state. Target objects for \textit{enveloping} and \textit{sucking} are respectively surrounded by red and green rectangles, which are the minimum-area bounding boxes.}
\end{figure*}

\textit{2) Orientation Optimization:} We present here a number of different scenarios generated by the proposed method that demonstrates the capability to optimize both enveloping and sucking orientations. Fig.~\ref{fig:enveloping} shows enveloping orientation optimization. It shows that our method can generate optimal orientations and opening distances for objects with different sizes and positions. A few examples of sucking-orientation optimization can be seen in Fig.~\ref{fig:sucking}. These experiments demonstrate the capability of our method to generate an optimal grasping orientation to secure collision-free grasping.

\begin{figure*}[htbp]
\centerline{\includegraphics[scale=0.16]{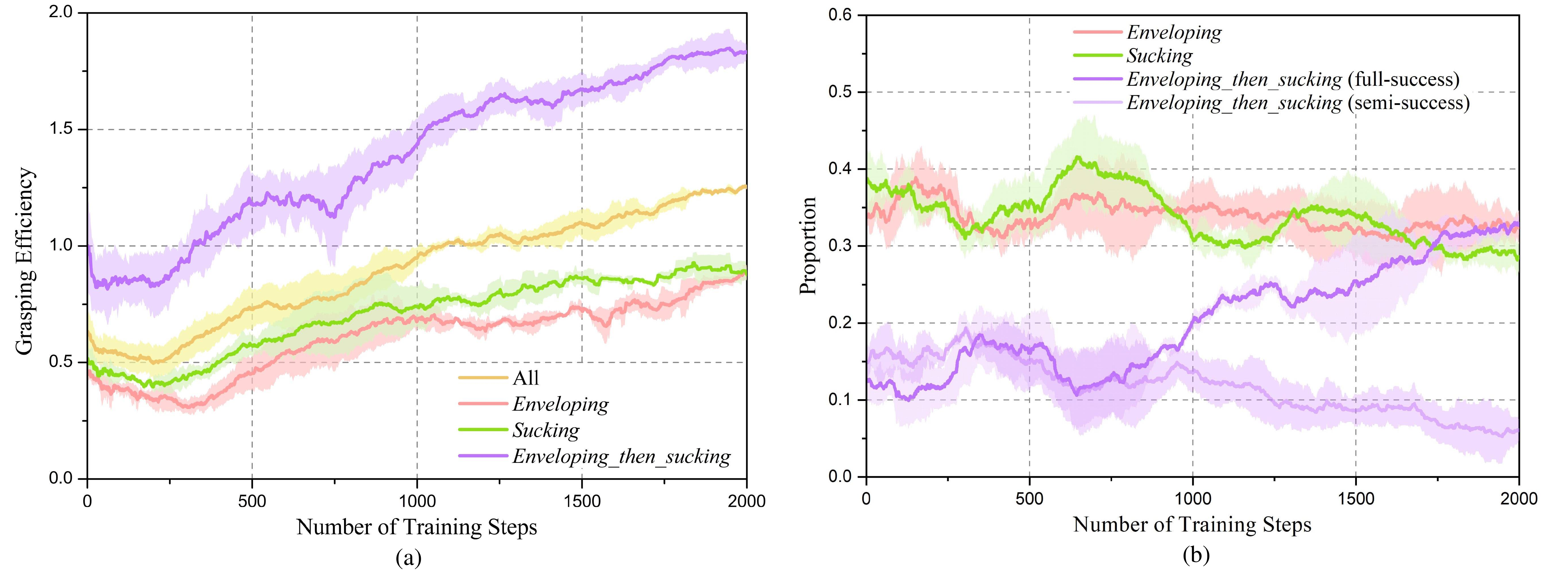}}
\caption{\label{fig:training} Training performance. (a) Grasping efficiencies of three types of actions over training steps. The grasping efficiencies increased significantly as the system picked the correct actions (i.e., the policy executes \textit{enveloping} and \textit{sucking} actions on objects that are suitable for enveloping and sucking, respectively, and maximizes the fully successful \textit{enveloping\_then\_sucking} actions). (b) Distributions of three types of actions among the successful actions over training steps. The \textit{enveloping\_then\_sucking} action has two success criteria (i.e., fully
successful action that picks up two objects and semisuccessful action that only picks up one object). The proportion of fully successful \textit{enveloping\_then\_sucking} actions increased significantly over training steps, contributing to the maximization of the grasping efficiency.}
\vspace{-0.3cm}
\end{figure*}

\subsection{Training}
We performed a series of simulations using the robot simulator CoppeliaSim. As described in Section~\ref{sec:gripper_design}, we simplified the SMG into a rigid-body model for the simulations. We imported the simplified gripper into the simulation platform and combined it with a UR5 robot for grasping. In order to model the real-world grasping environment, we simulated the simplified gripper and 3D objects dynamically using the Vortex physics engine in the simulations. All the simulations were repeated three times.

\textit{1) Evaluation Metrics:} We consider both the grasp success rate and the grasping efficiency in evaluating the multimodal grasping performance. We define an \textit{enveloping} or \textit{sucking} action as a successful action if one object is successfully picked up in this action. In addition, we label an \textit{enveloping\_then\_sucking} action as a fully successful action if two objects are lifted (i.e., both enveloping and sucking succeed) and as a semi-successful action if only one object is picked up (i.e., only enveloping or sucking succeeds). Any one of the three actions is marked as having failed if no object is successfully picked up. We describe grasping efficiency as the ratio of the number of objects successfully picked up to the number of actions performed. The value of the grasping efficiency can be greater than $100\%$, since two objects can be picked up simultaneously in one fully successful \textit{enveloping\_then\_sucking} action. We evaluate the two kinds of metrics both in training and in testing to verify our multimodal grasping policy.

\textit{2) Training:} In each trial during training, from one to five objects are randomly selected from the two grasp types and placed in the workspace. The total number of objects placed in the workspace, therefore, varies from two to $10$. For each training episode, the gripper iteratively performs grasping actions until either no objects remain on the table or the number of attempts reaches a maximum. In the simulations, the objects are first chosen randomly from the 3D model database, and they are then placed in the workspace with random positions and orientations.

Curves giving the variation of the average grasping efficiencies (i.e., the average number of items picked up per action), with standard deviations from three repetitions, are shown in Fig.~\ref{fig:training}(a). The results show that the grasping efficiencies increased significantly in $2000$ training steps. This occurs because the policy executes \textit{enveloping} and \textit{sucking} actions on objects that are suitable for enveloping and sucking, respectively, and maximizes the fully successful \textit{enveloping\_then\_sucking} actions). The method exhibits grasping efficiency of above $88\%$ on both \textit{enveloping} and \textit{sucking}, and above $183\%$ on \textit{enveloping\_then\_sucking}, all of which contribute to a grasping efficiency greater than $125\%$ on all the three grasping actions. Therefore, the results demonstrate that our DRL model is capable of learning effective multimodal grasping policies to optimize the efficiency of object grasping. Moreover, Fig.~\ref{fig:training}(b) depicts the distributions of four types of actions among the successful actions over training steps. The results show that there is an obvious increase in the proportion of fully successful \textit{enveloping\_then\_sucking} actions in later training steps, which can be attributed to our DRL method optimizing the grasping efficiency (i.e., minimizing the number of grasping actions). More details are described in the following sections.

\subsection{Baseline Comparisons and Ablation Studies}
For performance comparison, we first consider three baselines:
\begin{enumerate}
\item \textbf{Reactive Enveloping and Sucking Policy (E+S Reactive)}~This baseline is a greedy deterministic policy that chooses an action whose immediate estimated reward is greatest and plays greedily with respect to the current state. This baseline uses two FCNs—$\varOmega_e$ and $\varOmega_s$, supervised with binary classification to infer immediate estimated rewards for \textit{enveloping} and \textit{sucking}.
\item \textbf{Reactive Enveloping, Sucking and Enveloping\_then\_Sucking Policy (E+S+ES Reactive)}~This baseline is also a greedy deterministic policy that chooses an action with the maximal immediate estimated reward. This baseline is an augmented version of \textbf{E+S Reactive} but uses three FCNs—$\varOmega_e$, $\varOmega_s$, and $\varOmega_{es}$, to infer immediate estimated rewards for \textit{enveloping}, \textit{sucking}, and \textit{enveloping\_then\_sucking}, respectively. Both of E+S Reactive and E+S+ES Reactive networks are trained only on the immediate outcome of the action.
\item \textbf{DRL Enveloping and Sucking Policy (E+S DRL)}~This baseline is a subversion of the DRL scheme (i.e., DDQN) that takes the action that maximizes the expected sum of future discounted rewards. This baseline uses two Q-networks to infer Q-values for  \textit{enveloping} and \textit{sucking}.
\end{enumerate}

\begin{figure}[htbp]
\centerline{\includegraphics[scale=0.2]{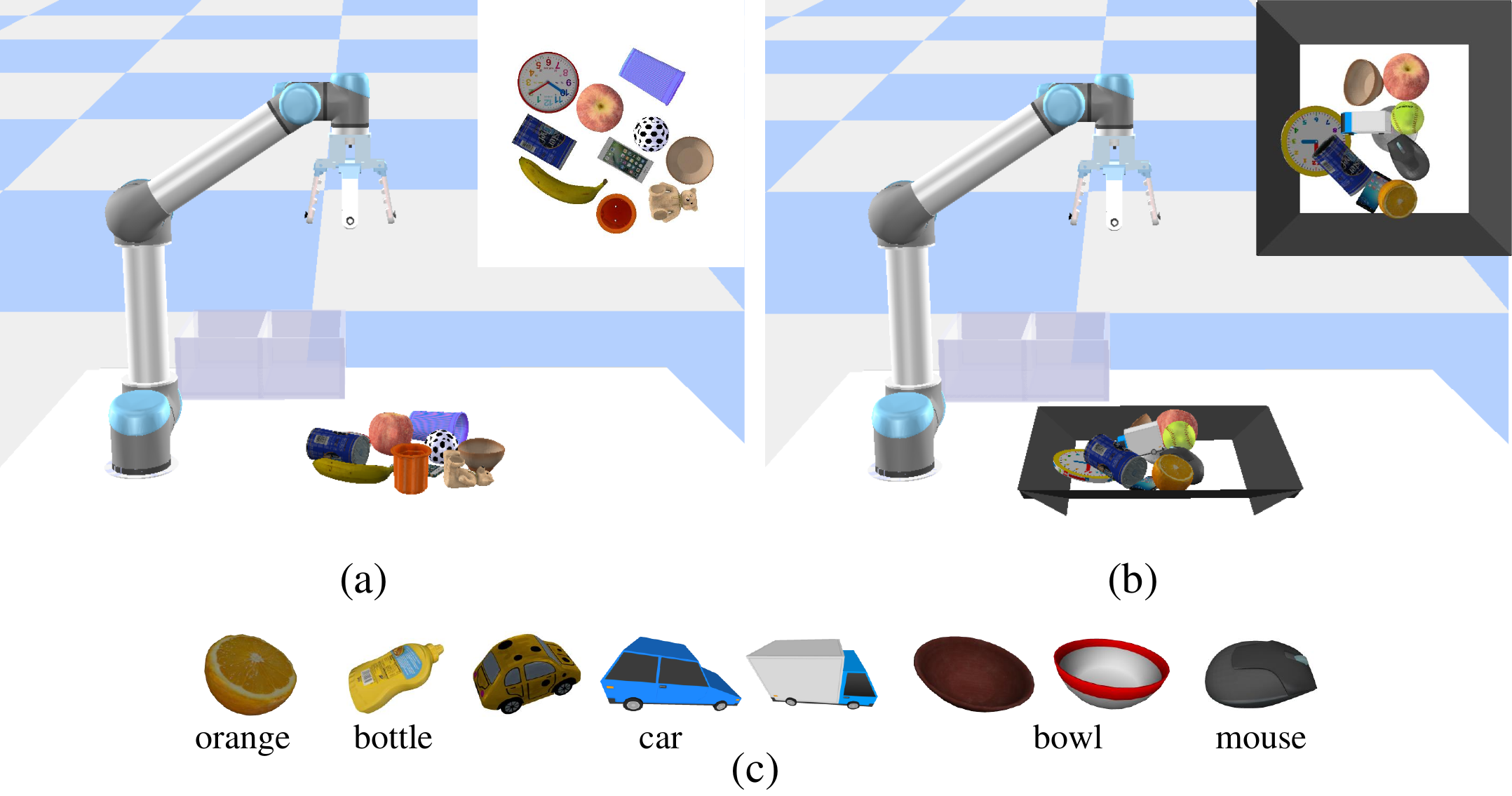}}
\caption{\label{fig:baseline_env} Grasping environments. (a) lightly cluttered. (b) Highly cluttered. (c) Objects that are suitable for both enveloping and sucking.}
\vspace{-0.3cm}
\end{figure}

We evaluated the performance of the DRL multimodal grasping policy (i.e., E+S+ES DRL(PE+OO) that executes actions with both pre-enveloping and orientation optimization) against the three baselines (please note that all of the three baselines execute actions with both pre-enveloping and orientation optimization) both in lightly and highly cluttered environments, which are shown in Fig.~\ref{fig:baseline_env}. In the relatively lightly cluttered scenarios, objects were placed randomly in the workspace and there were few objects piled together. In the highly cluttered scenarios, objects were closely placed and stacked in a small space. Each group of experiments contained $200$ executed actions and was repeated three times.

The results are shown in Table~\ref{tab:baseline}. There is little difference in success rates between the Reactive and DRL methods. The grasping efficiencies of E+S Reactive and E+S DRL are the same with their success rates since they are single-grasping-mode methods. However, the results suggest that both E+S+ES Reactive and E+S+ES DRL(PE+OO) achieve grasping efficiencies above $100\%$ in the lightly cluttered environment, which demonstrates the contribution of the multimodal grasping mode (i.e., \textit{enveloping\_then\_sucking} action). Although E+S+ES DRL(PE+OO) does not achieve the highest success rate against three baselines in lightly cluttered environments, it exhibits the highest grasping efficiencies with reliabilities of $119.5\%$ and $107.3\%$ in lightly and highly cluttered scenarios, respectively. E+S+ES DRL(PE+OO) outperforms E+S+ES Reactive in grasping efficiency due to the ability to plan long-term strategies for multimodal grasping. As shown in Fig.~\ref{fig:baseline_env}(c), some rounded objects with flat surfaces (oranges, bottles, cars, etc.) are suitable for both enveloping and sucking. The E+S+ES DRL(PE+OO) policy can execute either enveloping or sucking on these objects with respect to different ratios of two types of objects in the workspace, to optimize the grasping efficiency. In addition, the E+S+ES DRL(PE+OO) also shows good performance in highly cluttered environments. However, the grasping efficiency of E+S+ES Reactive decreases below $100\%$. Therefore, our DRL multimodal grasping framework can generate long-horizon sequential optimal grasping strategies to minimize the number of grasping actions (i.e., maximize the efficiency of object grasping).

\begin{table}[htbp]\small
\setlength{\tabcolsep}{0.5pt}
\renewcommand\arraystretch{1.0}
    \caption{\label{tab:baseline} Grasping performance of the DRL multimodal grasping policy and baselines (Mean with standard errors (SE) \%)}
    \begin{center}
    \footnotesize
        \begin{threeparttable}
        \begin{tabular}{lccccc}
             \toprule
             Method & \multicolumn{2}{c}{Lightly Cluttered}  & & \multicolumn{2}{c}{Highly Cluttered}    \\ 
            \cline{2-3} \cline{5-6} 
            &                                       $\zeta$             & $\eta$            &       & $\zeta$           &$\eta$    \\    \hline
            E+S Reactive                            &$92.3\pm 1.0$      & $92.3\pm 1.0$     &       & $90.0\pm 2.9$     & $90.0\pm 2.9$   \\       
            E+S+ES Reactive                         & \scriptsize{$\bm{92.7\pm 2.6}$}     & $109.8\pm 4.1$    &       & $89.7\pm 2.4$     & $99.3\pm 2.2$  \\         
            E+S DRL                                 & $91.2\pm 2.5$     & $91.2\pm 2.5$     &       & $89.5\pm 3.5$     & $89.5\pm 3.5$  \\
            E+S+ES DRL                              & $89.7\pm 4.6$     & $108.3\pm 6.2$    &       & $86.7\pm 2.9$     & $97.2\pm 4.6$  \\
            E+S+ES DRL(PE)                          & $90.5\pm 2.5$     & $113.7\pm 3.9$    &       & $86.8\pm 2.1$     & $99.8\pm 2.6$  \\
            E+S+ES DRL(OO)                          & $91.0\pm 1.5$     & $112.5\pm 5.5$    &       & $88.3\pm 3.8$     & $98.2\pm 3.8$  \\
            E+S+ES DRL(PE+OO)                       & $92.2\pm 1.0$     & \scriptsize{$\bm{119.5\pm 11.1}$}   &       & \scriptsize{$\bm{90.2\pm 2.0}$}     & \scriptsize{$\bm{107.3\pm 5.7}$}    \\
            \bottomrule
            \end{tabular}
            Depicted in Fig.~\ref{fig:baseline}. $\zeta$ and $\eta$ denote the success rate and grasping efficiency, respectively.\\
        \end{threeparttable}
        \label{bs}
    \end{center}
\vspace{-0.3cm}
\end{table}

\begin{figure}[htbp]
\centerline{\includegraphics[scale=0.16]{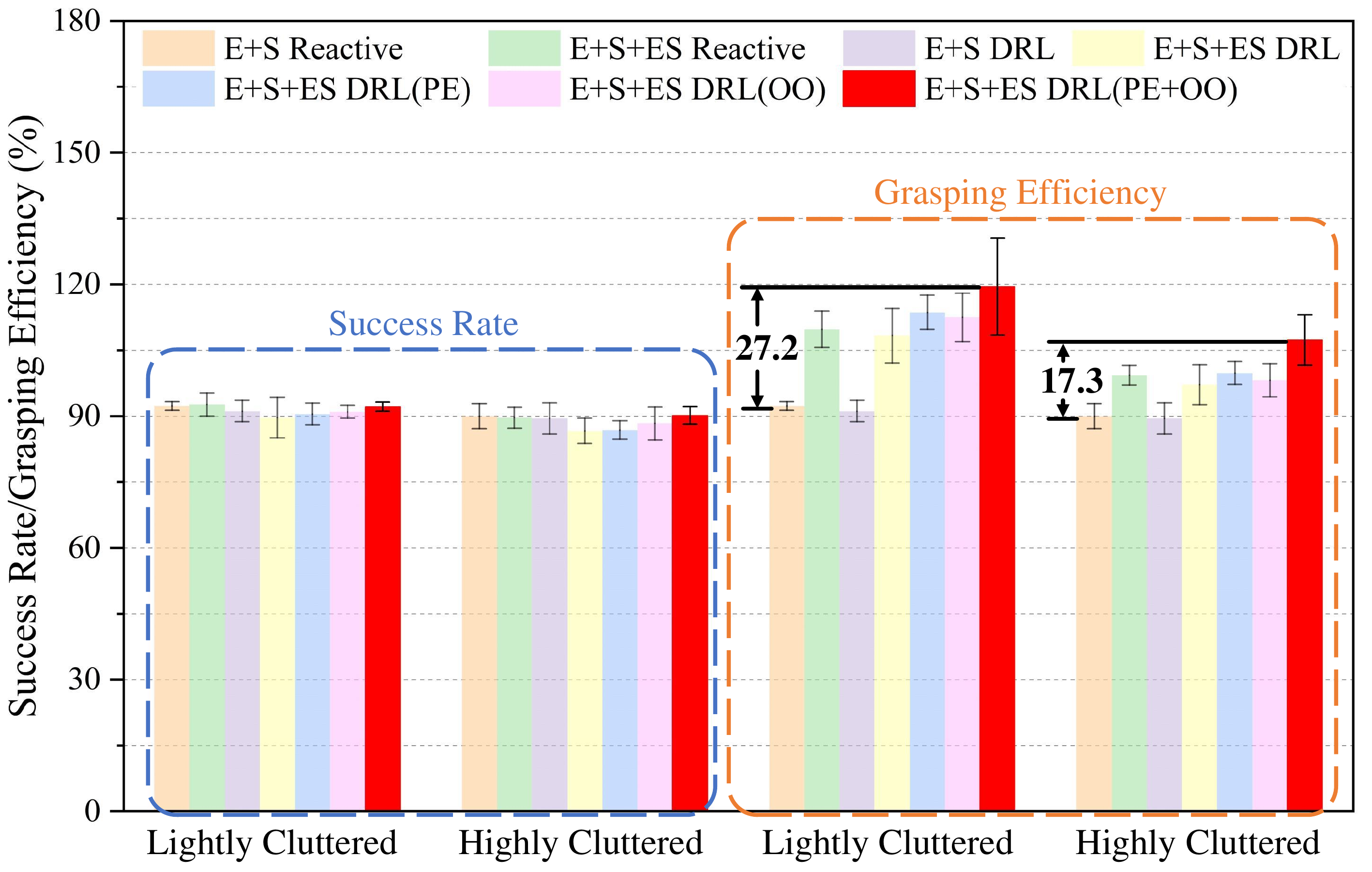}}
\caption{\label{fig:baseline}Performance of the baseline and ablation methods both in lightly and highly cluttered environments. E+S+ES DRL(PE+OO) executes more fully successful \textit{enveloping\_then\_sucking} actions than E+S+ES Reactive and other DRL multimodal grasping baselines that execute actions without pre-enveloping or orientation optimization, contributing to the maximization of the grasping efficiency. The grasping efficiency of E+S+ES DRL(PE+OO) is on average $27.2\%$ and $17.3\%$ higher than that of E+S Reactive in the lightly and highly cluttered environments, respectively. Error bars indicate standard errors (SE).}
\vspace{-0.3cm}
\end{figure}

\begin{figure*}[htbp]
\centerline{\includegraphics[scale=0.13]{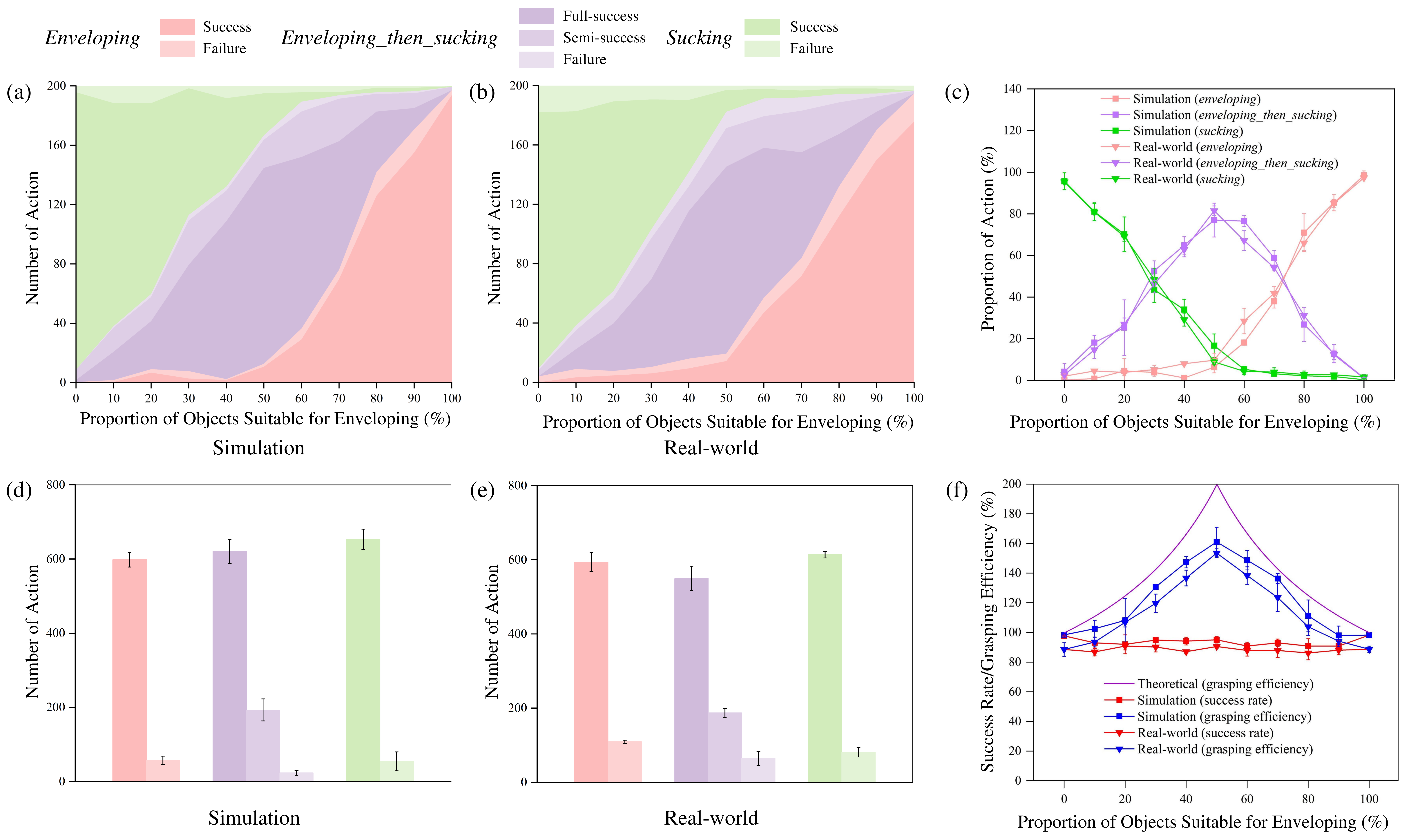}}
\caption{\label{fig:testing} Performance of the multimodal grasping policy both in simulations and in real-world tests. (a)–(c) Distributions of the three types of actions for $11$ different proportions of objects suitable for enveloping. Each proportion corresponds to one group experiment that contains $200$ actions and was repeated three times. Panels (d) and (e) show the distributions of the three types of actions in the total experiments, which contain $11$ groups of sub-experiments ($11\times200$ actions) and were repeated three times. (f) The success rate and grasping efficiency for different proportions of objects suitable for enveloping.}
\vspace{-0.2cm}
\end{figure*}

\begin{table*}[htbp]
\renewcommand\arraystretch{1}
\vspace{-0.2cm}
    \caption{\label{tab:testing-c} Distributions of the three types of actions for different proportions of objects suitable for enveloping (Mean \%)}
    \begin{center}
    \footnotesize
        \begin{threeparttable}
        \begin{tabular}{llllllllllll}
             \toprule
             Action & \multicolumn{11}{c}{Proportion of Objects Suitable For Enveloping (\%)}    \\ 
             \cline{2-12}
            &                                       $0$  &$10$     &$20$     &$30$     &$40$     &$50$     &$60$     &$70$     &$80$     &$90$     &$100$    \\    \hline
            Simulation (\textit{enveloping})                &$0.2$        &$0.8$   &$4.5$   &$3.8$   &$1.2$   &$6.3$   &$18.2$  &$38.0$  &$71.0$  &$85.3$  &$\bm{98.5}$  \\       
            Simulation (\textit{sucking})                   &$\bm{95.7}$       &$81.0$  &$70.2$  &$43.5$  &$34.0$  &$16.7$  &$5.3$   &$3.2$   &$2.2$   &$1.8$   &$0.3$   \\         
            Simulation (\textit{enveloping\_then\_sucking}) &$4.1$        &$18.2$  &$25.3$  &$52.7$  &$64.8$  &$\bm{77.0}$  &$76.5$  &$58.8$  &$26.8$  &$12.9$  &$1.2$   \\
            Real-world (\textit{enveloping})                &$2.0$        &$4.5$   &$3.8$   &$5.2$   &$8.0$   &$9.7$   &$28.5$  &$41.8$  &$66.0$  &$85.0$  &$\bm{97.3}$   \\
            Real-world (\textit{sucking})                   &$\bm{95.3}$       &$80.8$  &$69.3$  &$48.5$  &$29.2$  &$8.8$  &$4.3$   &$4.0$   &$2.8$   &$2.7$   &$1.7$   \\
            Real-world (\textit{enveloping\_then\_sucking}) &$2.7$        &$14.7$  &$27.0$  &$46.3$  &$62.8$  &$\bm{81.5}$  &$67.2$  &$54.2$  &$31.2$  &$12.3$  &$1.0$   \\
            \bottomrule
            \end{tabular}
            Depicted in Fig.~\ref{fig:testing}(c). The proportions of \textit{enveloping}, \textit{sucking}, and \textit{enveloping\_then\_sucking} actions reach maximum when $P_e$ are $100\%$, $0\%$, and $50\%$, respectively.
        \end{threeparttable}
        \label{bs}
    \end{center}
\end{table*}

\begin{table*}[htbp]\small
\renewcommand\arraystretch{1}
    \caption{\label{tab:testing-de} The distributions of the three types of actions in the total experiment (Mean)}
    \begin{center}
    \footnotesize
        \begin{threeparttable}
        \begin{tabular}{lccccccccc}
             \toprule
             Action & \multicolumn{2}{c}{\textit{Enveloping}}  & & \multicolumn{3}{c}{\textit{Enveloping\_then\_sucking}} & & \multicolumn{2}{c}{\textit{Sucking}}     \\ 
            \cline{2-3} \cline{5-7} \cline{9-10}
            &   success & failure   &   & full-success  & semi-success & failure & &success  & failure  \\  \hline
            Simulation  & $598.7$ & $57.0$  &   & $620.0$ & $193.0$ &$23.7$   &   & $653.3$ &$54.3$   \\       
            Real-world  & $594.0$ & $109.7$ &   & $549.7$ & $187.3$ &$64.7$   &   & $613.6$ & $81.0$   \\ 
            \bottomrule
            \end{tabular}
            Depicted in Figs.~\ref{fig:testing}(d) and (e).
        \end{threeparttable}
        \label{bs}
    \end{center}
\vspace{-0.1cm}
\end{table*}

\begin{table*}[htbp]\small
\renewcommand\arraystretch{1}
\vspace{-0.2cm}
    \caption{\label{tab:testing-f} The success rate and grasping efficiency for different proportions of objects suitable for enveloping (Mean \%)}
    \begin{center}
    \footnotesize
        \begin{threeparttable}
        \begin{tabular}{llllllllllll}
             \toprule
              & \multicolumn{11}{c}{Proportion of Objects Suitable For Enveloping (\%)}    \\ 
             \cline{2-12}
            &                                   $0$           &$10$     &$20$     &$30$     &$40$     &$50$     &$60$     &$70$     &$80$     &$90$     &$100$    \\    \hline
            Theoretical (grasping efficiency)   &$100.0$        &$111.1$   &$125.0$   &$142.9$   &$166.7$   &$\bm{200.0}$   &$166.7$  &$1.42.9$  &$125.0$  &$111.1$  &$100.0$  \\       
            Simulation (success rate)           &$97.7$       &$93.0$  &$92.0$  &$94.8$  &$94.2$  &$95.0$  &$90.8$   &$93.0$  &$90.8$   &$90.8$   &$98.2$   \\         
            Simulation (grasping efficiency)    &$97.7$      &$102.5$  &$108.2$  &$130.7$  &$147.3$  &$\bm{161.0}$  &$148.7$  &$136.3$  &$111.2$  &$98.0$  &$98.2$   \\
            Real-world (success rate)           &$88.5$        &$86.8$   &$90.8$   &$90.2$   &$87.0$   &$90.5$   &$87.8$  &$87.8$  &$86.2$  &$88.0$  &$88.7$   \\
            Real-world (grasping efficiency)    &$88.5$       &$93.5$  &$106.8$  &$119.7$  &$136.7$  &$\bm{153.5}$  &$138.3$   &$123.5$   &$103.8$   &$94.2$   &$88.7$   \\
            \bottomrule
            \end{tabular}
            Depicted in Fig.~\ref{fig:testing}(f). Grasping efficiencies reach maximum when $P_e$ is $50\%$.
        \end{threeparttable}
        \label{bs}
    \end{center}
\end{table*}

We conducted additional experiments to investigate the effect of pre-enveloping and orientation optimization. We consider three ablation baselines: \textbf{E+S+ES DRL} that executes actions without either pre-enveloping or orientation optimization, \textbf{E+S+ES DRL(PE)} that executes actions with only pre-enveloping, and \textbf{E+S+ES DRL(OO)} that executes actions with only orientation optimization. The results are shown in Table~\ref{tab:baseline}. We observed that E+S+ES DRL shows the worst performance among the four DRL multimodal grasping methods (i.e., E+S+ES DRL, E+S+ES DRL(PE), E+S+ES DRL(OO), and E+S+ES DRL(PE+OO)). This occurs because there are collisions between the fingers and elongated target objects (e.g., bananas, bottles, and remotes), and thus contribute to unsuccessful enveloping actions [see Fig.~\ref{fig:baseline}]. The grasping performance of E+S+ES DRL(OO) can be slightly improved with orientation optimization. However, its grasping efficiency remains much lower than that of E+S+ES DRL(PE+OO) since the collisions between the fingers and nontarget objects will sometimes lead to unsuccessful suction of the \textit{enveloping\_then\_sucking} actions. In addition, despite E+S+ES DRL(PE) executing actions with pre-enveloping, the collision-free grasps cannot be secured because of a lack of orientation optimization, also leading to both a lower success rate and a grasping efficiency than that of  E+S+ES DRL(PE+OO). These experiments indicate that pre-enveloping and orientation optimization are both necessary for the DRL scheme to achieve a high success rate and grasping efficiency.

\subsection{Performance Evaluation}
To evaluate the performance of our trained hybrid grasping model, we tested the method in scenarios with different ratios of the two types of objects to measure both the grasp rate and the grasping efficiency.

We define $P_e$ ($0\leq P_e\leq 1$) as the proportion of objects suitable for enveloping among all the objects. To test the effectiveness of our training policy, we performed $11$ groups of experiments with different values of $P_e$, varying from $0\%$ to $100\%$, as shown in Fig.~\ref{fig:testing}. Each group of experiments contained $200$ executed actions and was repeated three times. For each episode, two types of objects were randomly selected from the dataset according to $P_e$ (e.g., $P_e$= $30\%$ means that $30\%$ objects suitable for enveloping and $70\%$ objects suitable for sucking were selected), and they were then placed randomly in the workspace. One episode does not end until all objects in the workspace have been successfully picked up.

Fig.~\ref{fig:testing}(a) shows the dependence of the number of each of the three types of successful or failed actions as functions of $P_e$ in the simulations. As $P_e$ increases, the number of successful \textit{enveloping} actions increases accordingly, while the number of successful \textit{sucking} actions decreases. However, the number of successful \textit{enveloping\_then\_sucking} actions is spindle-shaped. It is smaller when $P_e$ approaches either $0\%$ or $100\%$. It increases as $P_e$ approaches $50\%$, and reaches a maximum when $P_e$ is $50\%$ (i.e., the ratio of the two types of objects is $1\colon1$) [see Table~\ref{tab:testing-c}]. This occurs because our multimodal grasping algorithm aims to maximize the expected sum of future rewards, and more fully successful \textit{enveloping\_then\_sucking} actions contribute to a larger sum of rewards. As shown in Fig.~\ref{fig:testing}(f), the grasping efficiency is always larger than the success rate, so long as the number of fully successful actions is greater than 0. We obtain the largest grasping efficiency $161.0\%$ [see Table~\ref{tab:testing-f}] in simulations containing the same number of objects suitable for enveloping and for sucking (i.e., $P_e=50\%$), outperforming other traditional single-grasping-mode methods, which only pick up one object in one action and have a grasping efficiency (equal to the success rate) less than $100\%$.

To evaluate the real-world performance and the generalizability of our proposed algorithm, we transferred the trained multimodal policy from the simulation to reality. In the real-world scenario, our SMG is mounted on the UR5 robot arm as an end-effector. An Intel RealSense D455 camera is mounted in the head to obtain visual and depth information about the workspace. It captures original RGB-D images with a resolution of $1280\times960$ pixels. 

For the simulations, we performed 11 groups of experiments in the real world for comparison, as shown in Fig.~\ref{fig:testing}. The correlations between the number of successes and failures for each of the three types of actions as functions of $P_e$ are similar to those from the simulation experiments. In all the experiments, the success rates remained above $85\%$ [see Table~\ref{tab:testing-f}]. The largest grasping efficiency was $153.5\%$ in reality [see Table~\ref{tab:testing-f}]. Table~\ref{tab:SuccessRate} lists the average success rates and grasping efficiencies for the three actions both in the simulations and in the real world. This demonstrates that our DRL policy trained in the simulations has good sim-to-real transfer performance.

\begin{table}[htbp]\small
    \caption{\label{tab:SuccessRate} Success Rate (Mean \%)}
    \begin{center}
    \footnotesize
        \begin{threeparttable}
        \begin{tabular}{lcccc}
            \toprule
            Action&$\mathcal{A}_{e}$&$\mathcal{A}_{s}$&$\mathcal{A}_{es}$&$\zeta$\\
            \midrule
            Simulation&$90.7$&$92.4$&$97.4$&$93.8$\\
            \midrule
            Real-world&$84.4$&$88.3$&$91.9$&$88.3$\\
            \bottomrule
            \end{tabular}
            $\zeta$ denotes the average success rate of all three grasping actions.
        \end{threeparttable}
        \label{bs}
    \end{center}
\end{table}

As discussed above, the grasping efficiency is strongly influenced by $P_e$. Let us consider an ideal case in which the grasp success rate is 1. Our aim is to minimize the number of grasping actions, and therefore as many fully successful \textit{enveloping\_then\_sucking} actions as possible should be executed. For example, if two objects in the workspace are suitable for enveloping and five for sucking (i.e., $P_e = 2/7$), we expect the robot to take two fully successful \textit{enveloping\_then\_sucking} actions and three successful \textit{sucking} actions to reach the maximum grasping efficiency of $7/5$. Thus, the grasping efficiency $\eta$ can be expressed as
\begin{equation}
\label{equ:gra_eff}
\eta =
\begin{cases}
\frac{1}{1-P_e},&{\text{if}}\ 0\leq P_e<0.5\\
\frac{1}{P_e},&{\text{otherwise.}}
\end{cases}
\end{equation}

The distribution of $\eta$ is shown in Table~\ref{tab:testing-f} (depicted in Fig.~\ref{fig:testing}(f)). Clearly, $\eta$ reaches a maximum at $P_e = 0.5$, which is in agreement with experimental results. If the values of $P_e$ are uniformly distributed on $(0, 1)$, then
\begin{equation}
\label{equ:expection}
\begin{aligned}
\begin{split}
& \mathbb{E}_{\eta}  = \int_{0}^{1}\eta dP_e = {\rm 2ln2}\\
& \mathbb{E}_{e} = \int_{0.5}^{1}(2P_e-1)dP_e = 0.25\\
& \mathbb{E}_{s} = \int_{0}^{0.5}(1-2P_e) dP_e = 0.25\\
& \mathbb{E}_{es} =\int_{0}^{0.5}P_e dP_e + \int_{0.5}^{1}(1-P_e)dP_e = 0.5
\end{split}
\end{aligned}
\end{equation}

where $\mathbb{E}_{\eta}$ denotes the expectation of the grasping efficiency $\eta$, and $\mathbb{E}_{e}$, $\mathbb{E}_{s}$, and $\mathbb{E}_{es}$ are the expectations for the proportions of \textit{enveloping}, \textit{sucking}, and \textit{enveloping\_then\_sucking} in all the grasping actions. Table~\ref{tab:GraEff} compares the average values from the experiments with the corresponding theoretical expectations. It shows that \textit{enveloping} and \textit{sucking} occur in almost equal proportion and that \textit{enveloping\_then\_sucking} accounts for the largest proportion among the three types of actions, whether theoretical or experimental. Furthermore, It shows that the grasping efficiency is on average $30.0\%$ and $28.4\%$ (equal to $(\eta-\zeta)/\zeta$) higher than the success rates in the simulations and in the real world, respectively, which demonstrates that our multimodal grasping framework significantly improves the grasping efficiency compared with one single grasping mode.

\begin{table}[htbp]\small
    \caption{\label{tab:GraEff} Experimental and theoretical results for the proportions of the three actions and the grasping efficiency (Mean \%)}
    \begin{center}
    \footnotesize
        \begin{tabular}{lcccc}
            \toprule
            Action&$\mathcal{A}_{e}$&$\mathcal{A}_{s}$&$\mathcal{A}_{es}$&$\eta$\\
            \midrule
            Simulation&$29.8$&$32.2$&$38.0$&$121.8$\\
            \midrule
            Real-world&$32.0$&$31.6$&$36.4$&$113.4$\\
            \midrule
            Theoretical&$25.0$&$25.0$&$50.0$&$138.2$\\
            \bottomrule
            \end{tabular}
        \label{bs}
    \end{center}
\vspace{-0.3cm}
\end{table}

\subsection{Limitations}
We observed the most frequent and notable failure cases due to our approach, as well as external factors.

Our method relies on object detection based on Mask R-CNN, which may result in grasping failure when the object is not detected. We selected a detection confidence threshold low enough to avoid the failure of object detection to the maximum extent possible. Besides, our DRL model takes depth images as input and grasping policies are highly related to the depth information of the object. However, we have noises on the depth image, especially on the edges of objects, when using the Intel Realsense D455 camera in real-world experiments. Wrong-depth information may contribute to nonoptimal grasping actions. We set the depth value to zero exceeding a given threshold to prevent noise from the Realsense camera.

\begin{figure}[htbp]
\centerline{\includegraphics[scale=0.2]{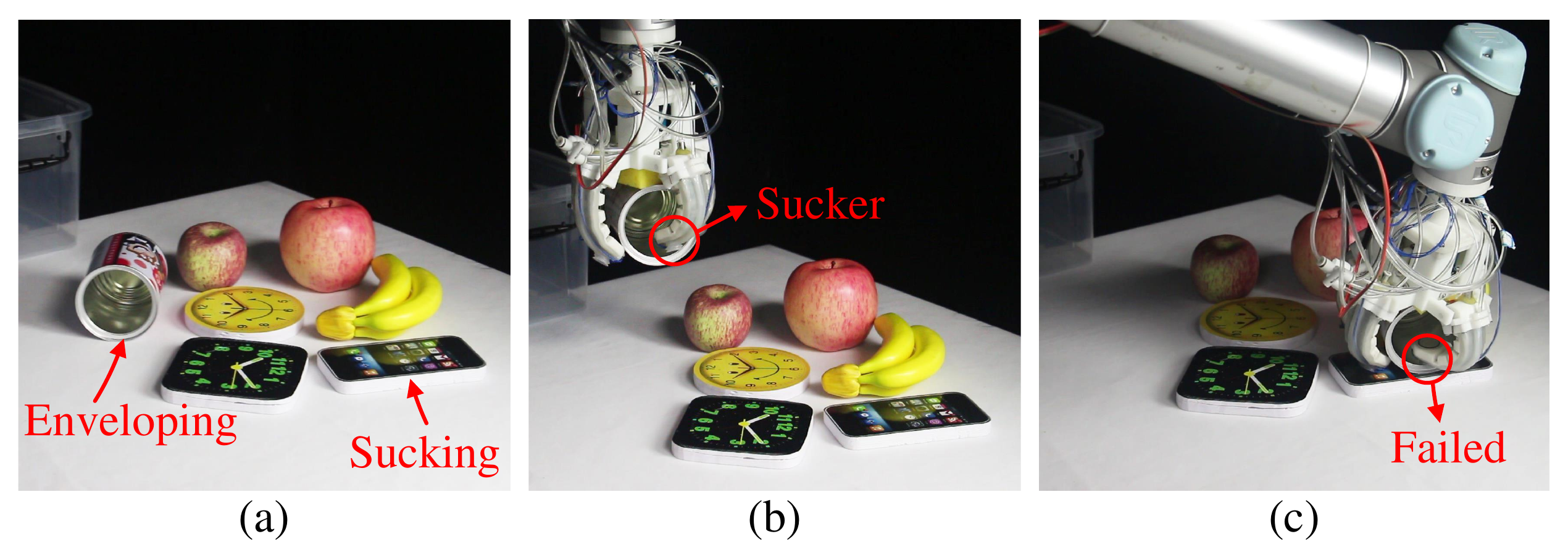}}
\caption{\label{fig:failed_action} Snapshots of a failed \textit{enveloping\_then\_sucking} action. (a) Target objects. (b) The sucker was blocked by the bottle after enveloping and before sucking. (c) Sucking failed.}
\vspace{-0.3cm}
\end{figure}

Some failure cases involved our soft gripper hardware. On the one hand, the \textit{enveloping\_then\_sucking} may fail to suck an object when the sucker used for sucking is blocked by another object, as shown in Fig.~\ref{fig:failed_action}. On the other hand, the learned model may execute failed \textit{enveloping} actions on objects that are close in size to the maximum allowed by the multimodal gripper. To solve this problem, we set the features of objects used in the simulation below the maximum size graspable by the gripper. Besides, we found failure cases caused by occasional collisions between the gripper and its environment. For example, the multimode action \textit{enveloping\_then\_sucking} fails to suck an object that moves too much due to perturbation in clutter. Moreover, despite the current version of the SMG is able to deal with objects with different shapes, it is limited to grasping this type of objects which exclude other type of daily life objects: articulated objects (e.g., scissors), flat objects with rough texture (e.g., keyboard), tiny objects (e.g., screw), etc.

\section{Conclusion}
\label{sec:conclusion}
In this work, we have developed a learning-based framework for robotic hybrid grasping, including gripper design, grasping modeling, simulation-based training, and sim-to-real transfer. Our aim is to minimize the number of grasping actions in order to optimize grasping efficiency. We designed an SMG that contains four fingers, with a vacuum cup on the back of each fingertip. It possesses the multimodal grasping ability, with three grasping modes—\textit{enveloping}, \textit{sucking}, and \textit{enveloping\_then\_sucking}—which enables the gripper to deal with heterogeneous objects and to grasp more than one object simultaneously. The proposed hybrid grasping learning method can achieve multistage autonomous grasping, and it can be used to explore the capacity of the SMG fully. We trained our learning model using DRL in simulations. We tested the trained model both in simulations and in reality. The results show that the distribution of the three executed actions is sensitive to the ratio of the two types of objects in the workspace. The number of \textit{enveloping} and \textit{sucking} actions executed are, respectively, positively and negatively correlated with $P_e$ (i.e., with the proportion of objects suitable for enveloping and  for sucking). However, the number of \textit{enveloping\_then\_sucking} actions executed increases only when $P_e$ approaches $50\%$, and it reaches a maximum at $P_e = 50\%$.

Our approach yielded a $93\%$ average success rate and a $161\%$ maximum grasping efficiency in the simulations, and it achieved an $88\%$ average success rate and a $154\%$ maximum grasping efficiency in real robot experiments. This demonstrates that our hybrid grasping model outperforms traditional single-grasping-mode methods, which have grasping efficiencies less than $100\%$. Furthermore, our DRL policy is able to deal with novel objects, and it can be reliably transferred from simulations to the real world. The code is available at \url{https://github.com/fukangl/SMG-multimodal-grasping}.

Future work includes the implementation of the proposed method to grasp a specified target in a cluttered environment; for example, optimizing the grasping actions in scenarios where the target object is surrounded by non-target objects with different features and overhead grasps are not allowed. We are also interested in reducing the size of the SMG to achieve fine manipulation and grasping tasks, such as assembly, surface finishing and shaping, etc.

{\appendix
In this Appendix, we determine the configurations of the gripper in the action space $\mathcal{A}$.

\subsection{Enveloping}
We define the following elementary rotations about the \textit{x}-, \textit{y}-, and \textit{z}-axes:

$$
\begin{aligned}
R_x(\alpha)=
\left[\begin{tabular}{ccc}
$1$ & 0 & $0$ \\
0 & $\operatorname{cos}(\alpha)$ & $-\operatorname{sin}(\alpha)$ \\ 
$0$ & $\operatorname{sin}(\alpha)$ & $\operatorname{cos}(\alpha)$ \\
\end{tabular}\right]
\end{aligned}
$$

$$
\begin{aligned}
R_y(\beta)=
\left[\begin{tabular}{ccc}
$\operatorname{cos}(\beta)$ & 0 & $\operatorname{sin}(\beta)$ \\
0 & 1 & 0\\
$-\operatorname{sin}(\beta)$ & 0 & $\operatorname{cos}(\beta)$ \\
\end{tabular}\right]
\end{aligned}
$$
and 
$$
\begin{aligned}
R_z(\gamma)=
\left[\begin{tabular}{ccc}
$\operatorname{cos}(\gamma)$ & $-\operatorname{sin}(\gamma)$ & 0\\ 
$\operatorname{sin}(\gamma)$ & $\operatorname{cos}(\gamma)$ & 0 \\
0 & 0 & 1 \\
\end{tabular}\right]
\end{aligned}
$$

The initial orientation of frame $G$ relative to the robot coordinates is $R_0=R_x(180^{\circ})$. Thus, after a rotation through the angle $\gamma_e$, the final orientation of frame $G$ relative to frame $W$ is $ R_e=R_0R_z(\gamma_e)$. Further, the center of each of the four fingertips attempts to move a given distance $\Delta$ below the target position $q_e$ when executing an enveloping action. In order to avoid collisions between the fingertips and the workspace, we define $\Delta = {\rm max\{5~cm}, q_e(3)\}$, where $q_e(3)$ denotes the $z$-coordinate of $q_e$. The coordinate of the target position relative to frame $G$ is then $q_{eg}=[0,0,h_p+h_f-\Delta]$. Therefore, $p_e=q_e-R_eq_{eg}$.

\subsection{Sucking}\label{A}

Algorithm 2 determines the sucking orientation $\alpha_s$, and the final orientation of frame $G$ relative to frame $W$ can be expressed as
\begin{equation}
\label{equ:R_s}
R_s =
\begin{cases}
R_0R_z(\gamma_s)R_x(-\theta_s),&{\text{if}}\ 45^{\circ}<\alpha_s\leq 135^{\circ}\\
R_0R_z(\gamma_s)R_y(\theta_s),&{\text{if}}\ 135^{\circ}<\alpha_s\leq 225^{\circ}\\
R_0R_z(\gamma_s)R_x(\theta_s),&{\text{if}}\ 225^{\circ}<\alpha_s\leq 315^{\circ}\\
R_0R_z(\gamma_s)R_y(-\theta_s),&{\text{otherwise.}}
\end{cases}
\end{equation}

\begin{algorithm}[H] 
\caption{Sucking Orientation Selection}  
\label{alg:A}
\hspace*{0.02in}{\bf Input:} 
set of obstacle factors $\{f_{oi}\}$, $\{\mathcal{F}_{oi}(\theta)\}$, $i=1,\dots,N$\\
\hspace*{0.02in}{\bf Output:}
sucking orientation $\alpha_s$
\begin{algorithmic}[1]
\State{$\mathcal{F}_{o}(\theta)\gets 1$ for $\theta\in[0^{\circ},360^{\circ})$, $\alpha_s\gets\varnothing$, initialize obstacle factor and sucking orientation}
\State{$\xi\gets 45^{\circ}$, initialize the threshold of obstacle region}
\While{$\alpha_s=\varnothing$}
    \If{${\rm min}\{{\mathcal{F}_{oi}(\theta)}\}\geq1$}
        \State{$\alpha_s\gets0^{\circ}$}
    \Else
        \For{$i=1:N$}
            \State{$\mathcal{F}_{o}(\theta)\gets \mathcal{F}_{o}(\theta)\cdot\mathcal{F}_{oi}(\theta)$}
        \EndFor
        \State{$\theta_{o}^{s1}\gets0^{\circ}$, $\theta_{o}^{e1}\gets0^{\circ}$, $\theta_{o}^{s2}\gets0^{\circ}$, $\theta_{o}^{e2}\gets0^{\circ}$, $\mathcal{F}_o^1\gets\mathcal{F}_{o}(0^{\circ})$}
        \For{$\theta = 0^{\circ}:359^{\circ}$}
            \If{$\mathcal{F}_{o}(\theta)\neq\mathcal{F}_o^1$}
                \State{$\theta_{o}^{e1}\gets\theta-1^{\circ}$}
                    \If{$\theta_{o}^{e1}-\theta_{o}^{s1}\geq\xi$ and $\mathcal{F}_{o}(\theta_{o}^{e1})\geq\mathcal{F}_o^1$}
                    \State{$\alpha_s\gets(\theta_{o}^{s1}+\theta_{o}^{e1})/2$, $\mathcal{F}_o^1=\mathcal{F}_{o}(\theta)$}
                    \EndIf
                \If{$\theta_{o}^{s1}=0^{\circ}$}
                    \State{$\theta_{o}^{e2}=\theta_{o}^{e1}$}
                \EndIf
                \State{$\theta_{o}^{s1}\gets\theta$}
            \EndIf
            \If{$\theta=359^{\circ}$}
                \If{ $\theta_{o}^{s1}\neq\theta$}
                    \State{$\theta_{o}^{e1}\gets\theta$, $\theta_{o}^{e2}\gets\theta_{o}^{s1}$}
                    
                    \If{$\theta_{o}^{e1}-\theta_{o}^{s1}\geq\xi$ and $\mathcal{F}_{o}(\theta)\geq\mathcal{F}_o^1$}
                    \State{$\alpha_s\gets(\theta_{o}^{s1}+\theta_{o}^{e1})/2$, $\mathcal{F}_o^1=\mathcal{F}_{o}(\theta)$}
                    \EndIf
                    
                \Else
                     \State{$\theta_{o}^{s2}\gets\theta$}
                \EndIf
                
            \EndIf
        \EndFor
        \If{$\mathcal{F}_{o}(0^{\circ})=\mathcal{F}_{o}(359^{\circ})$ and $360^{\circ}+\theta_{o}^{e2}-\theta_{o}^{s2}\geq\xi$ and $\mathcal{F}_{o}(0^{\circ})\geq\mathcal{F}_{o}^1$}
            \If{$\theta_{o}^{s2}+\theta_{o}^{e2}\geq\ 360^{\circ}$}
                \State{$\alpha_s\gets(\theta_{o}^{s2}+\theta_{o}^{e2})/2-180^{\circ}$}
            \Else
                \State{$\alpha_s\gets(\theta_{o}^{s2}+\theta_{o}^{e2})/2+180^{\circ}$}
            \EndIf
        \EndIf
    \EndIf
    \If{$\alpha_s=\varnothing$}
        \State{$i^{\ast}\gets\arg\max_{i}\{f_{oi}$, where $f_{oi}\neq1\}$}
        \State{$f_{oi^\ast}\gets1$}
    \EndIf
\EndWhile
\State{\bf{return:} $\alpha_s$}
\end{algorithmic}  
\end{algorithm}

The relationship between $\gamma_s$ and $\alpha_s$ is shown in Fig.~\ref{fig:action}. After a given bending angle $\theta_f$ is applied to each of the four fingers, the target location for the center of the sucker in frame $G$ can be specified as

\begin{equation}
\label{equ:q_sg}
q_{sg} =
\begin{cases}
[0, -d_s, h_p+h_s],&{\text{if}}\ 45^{\circ}<\alpha_s\leq 135^{\circ}\\
[-d_s, 0, h_p+h_s],&{\text{if}}\ 135^{\circ}<\alpha_s\leq 225^{\circ}\\
[0, d_s, h_p+h_s],&{\text{if}}\ 225^{\circ}<\alpha_s\leq 315^{\circ}\\
[d_s, 0, h_p+h_s],&{\text{otherwise}}
\end{cases}
\end{equation}

\noindent where $d_s$, $h_p$, and $h_s$ can be obtained from (\ref{equ:xs}). We then get $p_s=q_s-R_sq_{sg}$.

\subsection{Enveloping\_then\_sucking}\label{A}

The configuration $T_{es}=(p_e,R_e)\cup(p_s,R_s)$ can be obtained similarly to the above analysis.

}

\section*{Acknowledgement}
The authors would like to thank Shulin Du and Hang Liu for helping build the experimental platform, and Yixiao Li for video recordings. We particularly thank the reviewers for their professional and constructive feedback and comments.


\begin{thebibliography}{00}
\bibliographystyle{IEEEtran}

\bibitem{b1} 
A. Billard and D. Kragic, “Trends and challenges in robot manipulation," \emph{Science}, vol. 364, no. 6446, 2019.

\bibitem{b2} 
A. Izadbakhsh and S. Khorashadizadeh, “Robust impedance control of robot manipulators using differential equations as universal approximator," \emph{Int. J. Control}, vol. 91, no. 10, pp. 2170--2186, Jun. 2017.

\bibitem{b3} 
V. Subramaniam, S. Jain, J. Agarwal, and P. Y. V. Alvarado, “Design and characterization of a hybrid soft gripper with active palm pose control," \emph{Int. J. Robot. Res.}, vol. 39, no. 14, pp. 1668--1685, 2020.

\bibitem{b4} 
D. Rus and M. T. Tolley, “Design, fabrication and control of soft robots,” \emph{Nature}, vol. 521, no. 7553, pp. 467–475, 2015.

\bibitem{b5} 
S. Jun, C. Vito, F. Dario and S. Herbert, “Soft robotic grippers,” \emph{Adv. Mater.}, vol. 30, no. 29, May 2018.

\bibitem{b6} 
Q. Hu, E. Dong and D. Sun, “Soft gripper design based on the integration of flat dry adhesive, soft actuator, and microspine," \emph{IEEE Trans. Robot.}, vol. 37, no. 4, pp. 1065--1080, 2021.

\bibitem{b7} 
C. Tawk, A. Gillett, M. in het Panhuis, G. M. Spinks and G. Alici, “A 3D-printed omni-purpose soft gripper," \emph{IEEE Trans. Robot.}, vol. 35, no. 5, pp. 1268--1275, Oct. 2019.

\bibitem{b8} 
Y. Li, Y. Chen, Y. Yang and Y. Wei, “Passive particle jamming and its stiffening of soft robotic grippers," \emph{IEEE Trans. Robot.}, vol. 33, no. 2, pp. 446--455, April 2017.

\bibitem{b9} 
Y. Cui, X. Liu, X. Dong, J. Zhou and H. Zhao, “Enhancing the universality of a pneumatic gripper via continuously adjustable initial grasp postures," \emph{IEEE Trans. Robot.}, vol. 37, no. 5, pp. 1604–1618, Oct. 2021.

\bibitem{b10} 
X. Liu, Y. Zhao, D. Geng, S. Chen, X. Tan and C. Cao, “Soft humanoid hands with large grasping force enabled by flexible hybrid pneumatic actuators," \emph{Soft Robot.}, vol. 8, no. 2, pp. 175--185, April 2021.

\bibitem{b11} 
B. Fang \emph{et al.}, “Multimode grasping soft gripper achieved by layer jamming structure and tendon-driven mechanism," \emph{Soft Robot.}, doi: 10.1089/soro.2020.0065, Jun. 2021.

\bibitem{b12} 
B. Wu, I. Akinola, A. Gupta, F. Xu, J. Varley, D. Watkins-Valls and P. K. Allen, “Generative attention learning: A general framework for high-performance multi-fingered grasping in clutter," \emph{Auton. Robots}, vol. 44, no. 6, pp. 971--990, 2020.

\bibitem{b13} 
J. Mahler \emph{et al.}, “Learning ambidextrous robot grasping policies," \emph{Sci. Robot.}, vol. 4, no. 26, 2019.

\bibitem{b14} 
R. Balasubramanian, L. Xu, P. D. Brook, J. R. Smith and Y. Matsuoka, “Physical human interactive guidance: Identifying grasping principles from human-planned grasps," \emph{IEEE Trans. Robot.}, vol. 28, no. 4, pp. 899--910, Aug. 2012.

\bibitem{b15} 
B. Kehoe, A. Matsukawa, S. Candido, J. Kuffner and K. Goldberg, “Cloud-based robot grasping with the google object recognition engine," in \emph{Proc. IEEE Int. Conf. Robot. Autom.}, pp. 4263–4270, 2013.

\bibitem{b16} 
J. Bohg, A. Morales, T. Asfour and D. Kragic, “Data-driven grasp synthesis—A survey,"  \emph{IEEE Trans. Robot.}, vol. 30, no. 2, pp. 289--309, April 2014.

\bibitem{b17} 
X. B. Peng, M. Andrychowicz,W. Zaremba, and P. Abbeel, “Sim-to-real transfer of robotic control with dynamics randomization," in \emph{Proc. IEEE Int. Conf. Robot. Autom.}, pp. 3803–3810, 2018.

\bibitem{b18} 
W. Zhao, J. P. Queralta, and T. Westerlund, “Sim-to-real transfer in deep reinforcement learning for robotics: A survey," in \emph{Proc. IEEE Symp. Comput. Intell}, pp. 737--744, Dec. 2020.

\bibitem{b19} 
X. Yan \emph{et al.}, “Data-efficient learning for sim-to-real robotic grasping using deep point cloud prediction networks," 2019, \emph{arXiv:1906.08989}.

\bibitem{b20} 
A. Zeng, S. Song, S. Welker, J. Lee, A. Rodriguez, and T. Funkhouser, “Learning synergies between pushing and grasping with self-supervised deep reinforcement learning," in \emph{Proc. IEEE/RSJ Int. Conf. Intell. Robots Syst.}, pp. 4238–4245, 2018.

\bibitem{b21} 
S. Hasegawa, K.Wada, S. Kitagawa, Y. Uchimi, K. Okada, and M. Inaba, “GraspFusion: Realizing complex motion by learning and fusing grasp modalities with instance segmentation," in \emph{Proc. IEEE Int. Conf. Robot. Autom.}, pp. 7235–7241, 2019.

\bibitem{b22} 
A. Zeng \emph{et al.}, “Robotic pick-and-place of novel objects in clutter with multi-affordance grasping and cross-domain image matching," in \emph{Proc. IEEE Int. Conf. Robot. Autom.}, pp. 3750–3757, 2018.

\bibitem{b23} 
K. He, G. Gkioxari, P. Dollár, and R. Girshick, “Mask R–CNN," in \emph{Proc. IEEE Conf. Comput. Vis.}, pp. 2980--2988, 2017.

\bibitem{b24} 
E. Rohmer, S. P. Singh, and M. Freese, “V-rep: A versatile and scalable robot simulation framework," \emph{ IEEE/RSJ Int. Conf. Intell. Robots Syst.}, pp. 1321--1326, 2013.

\bibitem{b25} 
MultiChoiceGripper, Festo Corporate, 2013. [Online]. Available: https://www.festo.com

\bibitem{b26} 
W. Crooks, G. Vukasin, M. O’Sullivan, W. Messner, and C. Rogers, “Finray effect inspired soft robotic gripper: From the robosoft grand challenge toward optimization,” \emph{ Frontiers Robot. AI}, vol. 3, no. 70, p. 70, 2016.

\bibitem{b27} 
M. Guo \emph{et al.}, “Design of parallel-jaw gripper tip surfaces for robust grasping," in \emph{Proc. IEEE Int. Conf. Robot. Autom.}, pp. 2831-2838, 2017.

\bibitem{b29} 
R. Mutlu, G. Alici, M. in het Panhuis, and G. Spinks, “3D printed flexure hinges for soft monolithic prosthetic fingers,” \emph{ Soft Robot.}, vol. 3, no. 3, pp. 120–133, Sep. 2016.

\bibitem{b30} 
M. Manti, T. Hassan, G. Passetti, N. D’Elia, C. Laschi, and M. Cianchetti, “A bioinspired soft robotic gripper for adaptable and effective grasping,” \emph{ Soft Robot.}, vol. 2, no. 3, pp. 107–116, Sep. 2015.

\bibitem{b31} 
S. Terryn, J. Brancart, D. Lefeber, G. Van Assche, and B. Vanderborght, “Self-healing soft pneumatic robots," \emph{ Sci. Robot.}, vol. 2, no. 9, p. eaan4268, 2017.

\bibitem{b32} 
R. Deimel and O. Brock, “A novel type of compliant and underactuated robotic hand for dexterous grasping," \emph{Int. J. Robot. Res.}, vol. 35, no. 1-3, pp. 161--185, 2016.

\bibitem{b33} 
R. Deimel and O. Brock, “A compliant hand based on a novel pneumatic actuator," in \emph{Proc. IEEE Int. Conf. Robot. Autom.}, pp. 2047–2053, 2013.

\bibitem{b34} 
Y. Hao \emph{et al.}, “A multimodal, enveloping soft gripper: Shape conformation, bioinspired adhesion, and expansion-driven suction," \emph{IEEE Trans. Robot.}, vol. 37, no. 2, pp. 350--362, April 2021.

\bibitem{b35} 
S. Jun, R. Samuel, S. Bryan, F. Dario, and S. Herbert, “Versatile soft grippers with intrinsic electroadhesion based on multifunctional polymer actuators," \emph{Adv. Mater.}, vol. 28, no. 2, pp. 231--238, 2016.

\bibitem{b36} 
J. Guo, K. Elgeneidy, C. Xiang, N. Lohse, L. Justham, and J.Rossiter, “Soft pneumatic grippers embedded with stretchable electroadhesion,” \emph{Smart Mater. Struct.}, vol. 27, no. 5, 2018.

\bibitem{b37} 
J. Shintake, H. Shea and D. Floreano, “Biomimetic underwater robots based on dielectric elastomer actuators," in \emph{Proc. IEEE/RSJ Int. Conf. Intell. Robots Syst.}, pp. 4957–4962, 2016.

\bibitem{b38} 
B. Marc, K. Karl, Z. Jörg, N. Ulrich, and L. Andreas, “Reversible bidirectional shape-memory polymers," \emph{Adv. Mater.}, vol. 25, no. 32, pp. 4466--4469, 2013.

\bibitem{b39} 
W.Wei and A. Sung-Hoon, “Shape memory alloy-based soft gripper with variable stiffness for compliant and effective grasping,” \emph{Soft Robot.}, vol. 4, no. 4, pp. 379–389, 2017.

\bibitem{b41} 
Y. Wei \emph{et al.}, “A novel, variable stiffness robotic gripper based on integrated soft actuating and particle jamming," \emph{Soft Robot.}, vol. 3, no. 3, pp. 134--143, 2016.

\bibitem{b42} 
M.Zhu, Y. Mori,T.Wakayama, A.Wada, and S.Kawamura, “A fully multimaterial three-dimensional printed soft gripper with variable stiffness for robust grasping," \emph{Soft Robot.}, vol. 6, no. 4, pp. 507--519, 2019.

\bibitem{b43} 
J. Shintake, B. Schubert, S. Rosset, H. Shea, and D. Floreano, “Variable stiffness actuator for soft robotics using dielectric elastomer and low-melting-point alloy," in \emph{Proc. IEEE/RSJ Int. Conf. Intell. Robots Syst.}, 1097–1102, 2015.

\bibitem{b44} 
OctopusGripper, Festo Corporate, 2017. [Online]. Available: https://www.festo.com

\bibitem{b45} 
I. Lenz, H. Lee, and A. Saxena, “Deep learning for detecting robotic grasps," \emph{Int. J. Robotics Res.}, vol. 34, no. 4-5, pp. 705--724, 2015.

\bibitem{b46} 
A. Saxena, J. Driemeyer, and A. Y. Ng, “Robotic grasping of novel objects using vision," \emph{Int. J. Robotics Res.}, vol. 27, no. 2, pp. 157--173, 2008.

\bibitem{b47} 
L. Pinto and A. Gupta, “Supersizing self-supervision: Learning to grasp from 50k tries and 700 robot hours,” in \emph{Proc. IEEE Int. Conf. Robot. Autom.}, pp. 3406–3413, May 2016.

\bibitem{b48} 
J. Cai, H. Cheng, Z. Zhang and J. Su, “MetaGrasp: Data efficient grasping by affordance interpreter network," in \emph{Proc. IEEE Int. Conf. Robot. Autom.}, pp. 4960–4966, 2019.

\bibitem{b50} 
X. Chen \emph{et al.}, “GRIP: Generative robust inference and perception for semantic robot manipulation in adversarial environments," in \emph{Proc. IEEE Int. Conf. Intell. Robots Syst.}, pp. 3988--3995, 2019.

\bibitem{b51} 
M. Danielczuk \emph{et al.}, “Segmenting unknown 3D objects from real depth images using mask r–cnn trained on synthetic data," in \emph{Proc. IEEE Int. Conf. Robot. Autom.}, pp. 7283--7290, 2019.

\bibitem{b52} 
J. Mahler \emph{et al.}, “Dex-Net 2.0: Deep learning to plan robust grasps with synthetic point clouds and analytic grasp metrics," in \emph{Proc. Int. Conf. Robot.: Sci. Syst.}, 2017.

\bibitem{b53} 
B. Wu, I. Akinola and P. K. Allen, “Pixel-attentive policy gradient for multi-fingered grasping in cluttered scenes," in \emph{Proc. IEEE/RSJ Int. Conf. Intell. Robots Syst.}, pp. 1789–1796, 2019.

\bibitem{b54} 
D. Morrison, P. Corke, and J. Leitner, “Closing the loop for robotic grasping: A real-time, generative grasp synthesis approach," in \emph{Proc. Robot.: Sci. Syst.}, 2018.

\bibitem{b55} 
F. -J. Chu, R. Xu and P. A. Vela, “Real-world multiobject, multigrasp detection,” \emph{IEEE Robot. Autom. Lett.}, vol. 3, no. 4, pp. 3355–3362, Oct. 2018.

\bibitem{b56} 
A. Zeng \emph{et al.}, “Multi-view self-supervised deep learning for 6d pose estimation in the amazon picking challenge,” \emph{ in Proc. IEEE Int. Conf. Robot. Autom.}, pp. 1386–1383, 2017.

\bibitem{b57} 
D. Kalashnikov \emph{et al.}, “Qt-opt: Scalable deep reinforcement learning for vision-based robotic manipulation,” 2018, \emph{arXiv:1806.10293}.

\bibitem{b58} 
D. Quillen, E. Jang, O. Nachum, C. Finn, J. Ibarz and S. Levine, “Deep reinforcement learning for vision-based robotic grasping: A simulated comparative evaluation of off-policy methods," in \emph{Proc. IEEE Int. Conf.
Robot. Autom.}, pp. 6284–6291, 2018. 

\bibitem{b59} 
L. Pinto, M. Andrychowicz, P. Welinder, W. Zaremba, and P. Abbeel, “Asymmetric actor critic for image-based robot learning, in "\emph{Proc. Robot.: Sci Syst.}, 2018.

\bibitem{b61} 
T. Kim, Y. Park, Y. Park, and I. H. Suh, “Acceleration of actor-critic deep reinforcement learning for visual grasping in clutter by state representation learning based on disentanglement of a raw input image,” 2020, \emph{arXiv:2002.11903}.

\bibitem{b62} 
S. Hasegawa, K. Wada, Y. Niitani, K. Okada and M. Inaba, “A three-fingered hand with a suction gripping system for picking various objects in cluttered narrow space," in \emph{Proc. IEEE Int. Conf. Intell. Robots Syst.}, pp. 1164–1171, 2017.

\bibitem{b63} 
K. Yamaguchi, Y. Hirata and K. Kosuge, “Development of robot hand with suction mechanism for robust and dexterous grasping,” in \emph{Proc. IEEE/RSJ Int. Conf. Intell. Robots Syst.}, pp. 5500–5505, 2013.

\bibitem{b64} 
V. Mnih \emph{et al.}, “Human-level control through deep reinforcement learning," \emph{Nature}, vol. 518, no. 7540, pp. 529–533, 2015.

\bibitem{b65} 
H. Van Hasselt, A. Guez, and D. Silver, “Deep reinforcement learning with double q-learning,"  in \emph{Proc. AAAI Conf. Artif. Intell.}, vol. 2, pp. 2095–2100, 2016.

\bibitem{b66} 
G. Huang, Z. Liu, K. Q. Weinberger, and L. van der Maaten, “Densely connected convolutional networks," in \emph{Proc. IEEE Conf. Comput. Vision Pattern Recognit.}, 2017.

\bibitem{b67} 
J. Deng, W. Dong, R. Socher, L.-J. Li, K. Li, and L. Fei-Fei, “Imagenet: A large-scale hierarchical image database," in \emph{Proc. IEEE Conf. Comput. Vis. Pattern Recognit..}, pp. 248–255, 2009.

\bibitem{b68} 
D. P. Kingma and J. Ba, “Adam: A method for stochastic optimization,” 2014, \emph{arXiv:1412.6980}.

\bibitem{b69} 
R. S. Sutton and A. G. Barto, “Reinforcement learning: An introduction," Cambridge, MA, USA: MIT Press, 2018.

\bibitem{b70} 
K. He, X. Zhang, S. Ren, and J. Sun, “Delving deep into rectifiers: Surpassing human-level performance on imagenet classification,” in \emph{Proc. IEEE Int. Conf. Comput. Vis.}, pp. 1026–1034, 2015.

\bibitem{b71} 
T.-Y. Lin \emph{et al.}, “Microsoft COCO: Common objects in context,” in \emph{Proc. Eur. Conf. Comput. Vis.}, pp. 740–755, 2014.

\bibitem{b72} 
Universal Robots, “Technical Specifications UR5,” accessed 2022-08-28. [Online]. Available: https://www.universal-robots.com/

\end{thebibliography}
\end{document}